\documentclass[10pt,twocolumn,letterpaper]{article}

\usepackage[pagenumbers]{cvpr} 

\usepackage[dvipsnames]{xcolor}

\usepackage{mathrsfs}
\usepackage{tikz}
\usetikzlibrary{arrows.meta}
\usepackage{overpic}
\usepackage{pict2e}
\usepackage{dsfont}
\usepackage[most]{tcolorbox}
\usepackage[outline]{contour}
\usepackage{standalone}
\usepackage[normalem]{ulem}
\usepackage{bm}

\usepackage{microtype}      
\usepackage{xspace}         
\usepackage{mathtools}      
\usepackage{nicefrac,xfrac} 
\usepackage{mathtools}

\usepackage{stmaryrd}  
\usepackage{newfloat}  
\usepackage{savesym}
\usepackage{algorithm}
\usepackage{algorithmicx}
\usepackage[noend]{algpseudocode} 
\algrenewcommand\textproc{}
\usepackage{graphicx}
\usepackage{grffile}  
\usepackage{soul}
\usepackage{wrapfig}
\usepackage{bbm}
\usepackage{flushend}
\usepackage{tabularx}
\usepackage{ctable}            
\usepackage[a]{esvect}           
\usepackage{multirow}

\usepackage{wrapfig}
\usepackage{textcomp}    
\usepackage{enumitem}    

\newcommand{\mymath}[2]{\newcommand{#1}{\TextOrMath{$#2$\xspace}{#2}}}
\mymath{\NumFrames}{F}
\mymath{\width}{W}
\mymath{\height}{H}
\mymath{\channel}{C}
\mymath{\inputvideo}{\bm{x}}
\mymath{\outputvideo}{\bm{\hat{y}}}
\mymath{\targetvideo}{\bm{y}}

\mymath{\warpmask}{\bm{M}}
\mymath{\latent}{\bm{z}}
\mymath{\forwardwarping}{\mathcal{F}}
\mymath{\disparityscaling}{S}
\mymath{\vaeencoder}{E}
\mymath{\vaedecoder}{D}
\mymath{\diffusionunet}{f}

\definecolor{cvprblue}{rgb}{0.21,0.49,0.74}
\usepackage[pagebackref,breaklinks,colorlinks,citecolor=cvprblue]{hyperref}
\def\paperID{9563} 
\def\confName{ICCV} 
\def\confYear{2025}

\title{
Restereo: Diffusion stereo video generation and restoration
}

\author{Xingchang Huang$^{1,2}$
\quad
Ashish Kumar Singh$^{4}$
\quad
Florian Dubost$^{4}$
\quad
Cristina Nader Vasconcelos$^{5}$
\\
Sakar Khattar$^{4}$
\quad
Liang Shi$^{4}$
\quad
Christian Theobalt$^{1,2}$
\quad
Cengiz {\"O}ztireli$^{3,4}$
\quad
Gurprit Singh$^{1,2}$
\\
\\
$^{1}$Max Planck Institute for Informatics
\quad 
$^{2}$VIA-Center Saarb{\"u}cken
\\
$^{3}$University of Cambridge
\quad 
$^{4}$Google
\quad
$^{5}$Google DeepMind
}

\begin{document}
\maketitle

\begin{abstract}
Stereo video generation has been gaining increasing attention with recent advancements in video diffusion models. 
However, most existing methods focus on generating 3D stereoscopic videos from monocular 2D videos. 
These approaches typically assume that the input monocular video is of high quality, making the task primarily about inpainting occluded regions in the warped video while preserving disoccluded areas.
In this paper, we introduce a new pipeline that not only generates stereo videos but also enhances both left-view and right-view videos consistently with a single model.
Our approach achieves this by fine-tuning the model on degraded data for restoration, as well as conditioning the model on warped masks for consistent stereo generation. 
As a result, our method can be fine-tuned on a relatively small synthetic stereo video datasets and applied to low-quality real-world videos, performing both stereo video generation and restoration.
Experiments demonstrate that our method outperforms existing approaches both qualitatively and quantitatively in stereo video generation from low-resolution inputs.
\end{abstract}

\section{Introduction}

Stereo video generation is becoming increasingly crucial for creating immersive experiences in modern VR applications.
Recently, diffusion models have been showing great potential for realistic video generation from text prompts.
However, most of the existing models focus on generating monocular videos, while stereo video generation remains
under-explored.

Training new diffusion models for stereo videos from scratch can be expensive and time-consuming.
There are recent progress on stereo video generation using training-free strategies by leveraging pretrained models such as Stable Diffusion~\cite{rombach2022high} and ModelScope/Zeroscope~\cite{wang2023modelscope}.
Existing training-free camera control methods (e.g., \citet{hou2024training}) are unsuitable for fine-grained stereo video generation.
\citet{shi2024stereocrafterzero} propose zero-shot stereo video generation with noisy restart.
Similar works include SVG from~\citet{dai2024svg}, StereoDiffusion from~\citet{wang2024stereodiffusion} and T-SVG from~\citet{jin2024t}.
These training-free methods do not require training on large datasets, making them lightweight and easy to use.
However, these methods are limited to some extend as they only rely on the pretrained diffusion models to inpaint the unknown region of the right view image in the latent space.
Since the pretrained diffusion models are not specifically trained on stereo data, the inpainting can be spatially and temporally inaccurate.
After decoding the inpainted latent representation, the generated right view images is not guaranteed to be consistent with the left view image.
This may also lead to artifacts around the occluded region in the right view images.

On the other hand, training-based methods such as StereoCrafter~\cite{zhao2024stereocrafter}, SpatialMe~\cite{zhang2024spatialme}, StereoConversion~\cite{mehl2024stereo} and ImmersePro~\cite{shi2024immersepro} have shown promising results in stereo video generation.
These methods are designed to train on a large-scale dataset of stereo videos from the internet or the movie industry, and have become foundation models in stereo video generation.
While theses datasets are diverse with indoor and outdoor scenes, they are expensive to capture and annotate for individuals and have not been publicly available.
A bigger issue for these methods is that they are trained to inpaint the left warped videos to generate the right videos by construction.
These models assume the input video is high-quality and learn to maintain details from the input view and inpaint the occluded regions naturally.
But when the input video is low-quality, their models still keep the details from the input view, making the generated stereo video low-quality and cannot manage to enhance the video at the same time.

In this work, we suggest a novel idea for simultaneous stereo video generation and restoration.
To achieve this, we propose to use data degradation (e.g., noise addition, blurring, compression) during training, inspired by Real-ESRGAN~\citep{wang2021real}.  
We design a consistent training and inference pipeline to learn simultaneously generation of both left and right views with restoration.
In this way, we can not only enhance the input left-view video, but also generating a right-view video consistent to the left-view one.
Further, this pipeline can be trained via a synthetic dataset that are relatively easier to generate compared to real-world data.
Synthetic data come with ground truth information like depth, which removes the complicated preprocessing pipeline including stereo matching, depth estimation, data filtering, etc as proposed in
StereoCrafter~\cite{zhao2024stereocrafter}.

In summary, we make the following contributions:
\begin{itemize}
    \item a consistent training and inference pipeline for both left and right views conditioned on warped masks,
    \item a process of data augmentation with image degradations to train the model for robust stereo video generation and restoration simultaneously,
    \item a novel pipeline for stereo video generation and restoration using synthetic data (both the pipeline and data will be publicly available upon acceptance).
    
\end{itemize}

\begin{figure*}[htb]
\includegraphics[width=1.0\textwidth]
{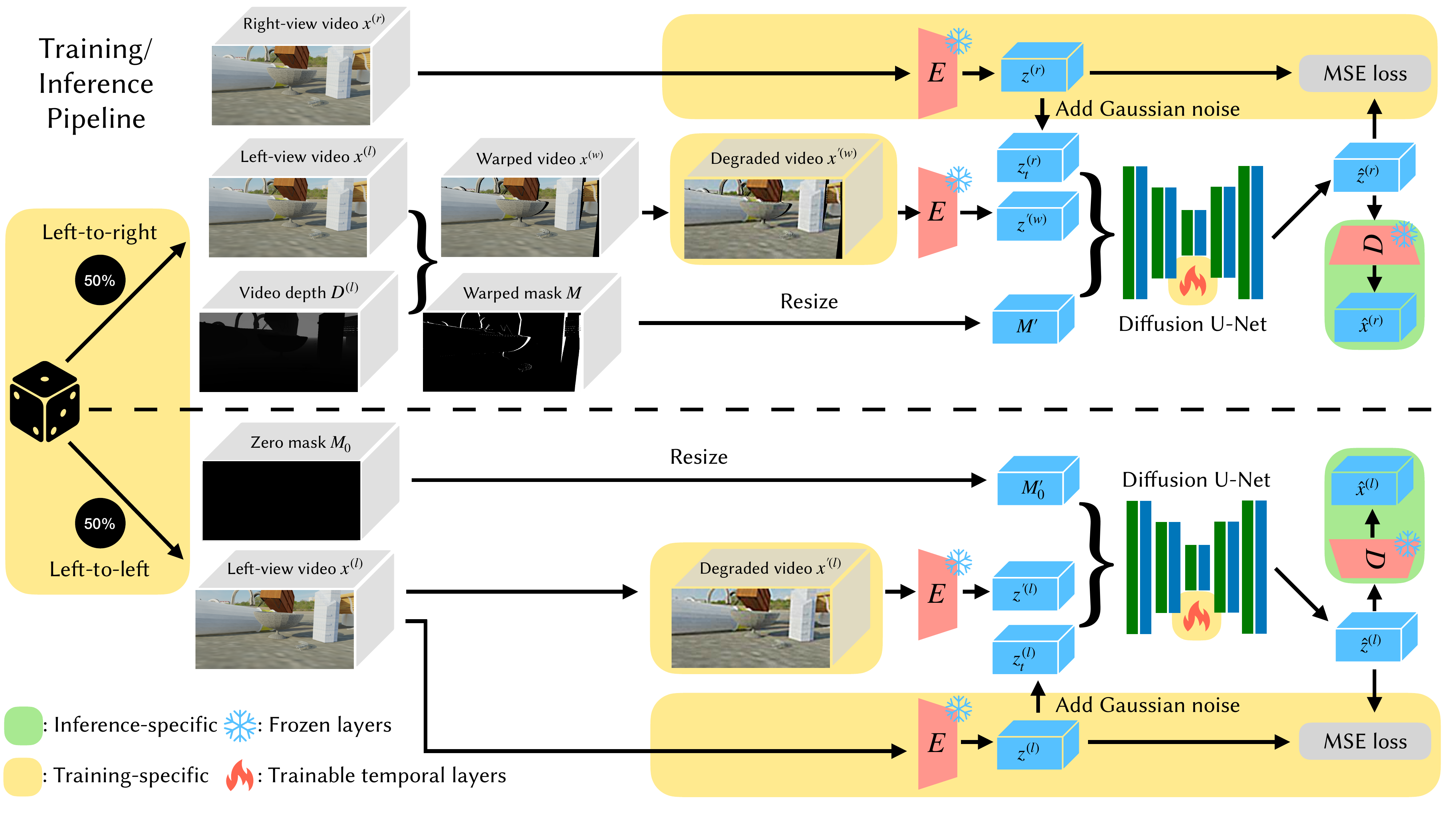}
\centering
\caption{
Training and inference pipeline of our method.
We fine-tune the Diffusion U-Net for both left-to-right and left-to-left generation and restoration branches.
During training, we randomly sample a branch, where left-to-right requires depth maps, forward warping and the right-view target video.
For left-to-left, no warping is required and we use a zero mask as the condition and left-view video as the target.
Both branches require data augmentation/degradation during training.
During inference, we run both branches as well as the decoder to generate videos for both views, without using the yellow boxes.
Note that the two U-Nets share the same weights and CLIP~\cite{radford2021learning} features of $\latent'^{(w)}$ and $\latent'^{(l)}$ are also part of the conditional input to the U-Net omitted for simplicity.
Details are discussed in~\Cref{sec:training,sec:inference}. 
}
\label{fig:pipeline}
\end{figure*}

\section{Related work}
Here we mainly focus on diffusion based methods for stereo image and video generation, as they represent current state-of-the-art quality.
As discussed in StereoCrafter~\cite{zhao2024stereocrafter}, methods relying on 3D representation for reconstruction like NeRF~\cite{mildenhall2021nerf}, 3DGS~\cite{kerbl20233d} are not the best choice for stereo generation.
This is these methods require both capturing a lot of frames for each scene and estimating the camera pose of each frame from the input videos.
It becomes challenging for these methods based on single-view inputs, with dynamic objects, or visual effects such as fog or fire.
While some work based on 3D representation such as DynIBaR~\cite{li2023dynibar}, RoDynRF~\cite{liu2023robust} can be robust to monocular inputs, they can still contain artifacts and without the capability to generate content.
As a result, we focus on recent diffusion-based method as practical solutions for producing stereoscopic videos.

\paragraph{Training-free diffusion-based stereo generation.}
\citet{wang2024stereodiffusion} propose StereoDiffusion, to generate stereo images via copying and shifting in the latent space of pretrained image diffusion models.
This method is training-free but requires inpainting in the latent space which can be error-prone and does not scale well to stereo video generation.
The RePaint algorithm proposed by \citet{lugmayr2022repaint} can be repurposed for stereo image generation, similar to the latent inpainting part in~\citet{wang2024stereodiffusion}.
\citet{dai2024svg} propose a training-free method for stereo video generation using pretrained video diffusion models.
Compared to~\citet{wang2024stereodiffusion}, this works better in terms of temporal consistency, but still rely on a pretrained model that is not trained for stereo video generation.
Although there exists off-the-shelf video inpainting tool~\cite{li2022towards,zhou2023propainter}, they might generate blurry content and inconsistent artifacts as they are not designed to be integrated into the pipeline.
Immersity AI~\cite{immersityai} and Owl3D~\cite{owl3d} are 2D-to-3D conversion softwares that can generate more consistent results.
However, all these methods are designed to maintain the details of the left-view input and assume the input is high-quality.

\paragraph{Training-based diffusion-based stereo generation.}
Recently proposed training-based methods show promising results with spatial and temporal consistency. StereoCrafter~\citep{zhao2024stereocrafter}, SpatialMe~\citep{zhang2024spatialme}, StereoConversion~\citep{mehl2024stereo}, ImmersePro~\citep{shi2024immersepro}.
But these methods often requires a large-scale dataset of stereo videos from the internet or movies, which is usually expensive to capture and annotate.
They may requires complicated pipeline for data preprocessing, stereo matching to obtain the video depth and warped video as the training data.
One example is shown in the StereoCrafter~\cite{zhao2024stereocrafter} where accurate stereo matching is crucial for subsequent steps.
We adopt a training-based strategy as well but with synthetic data, as they are easier to obtain without complicated data preprocessing. 
Similarly, these models are trained to preserve the details of input videos, without considering the input can be low-quality.

\paragraph{Multi-view and video diffusion models.}
Recent progress on multi-view and video diffusion models has shown great capability in novel view synthesis.
CAT3D~\cite{gao2024cat3d} supports novel view synthesis from single- or multi-view images combing multi-view diffusion and NeRFs.
\citet{xie2024sv4d} extends Stable Video Diffusion (SVD)~\cite{blattmann2023stable} to Stable Video 4D (SV4D), which can reconstruct a 4D scene from a single input video.
But their method only considers a foreground animated object without background.
Similar work includes Generative Camera Dolly~\cite{van2024generative} and CAT4D~\cite{wu2024cat4d}.
But these works are focused on novel view synthesis from large camera baselines and not directly usable for stereo video generation.

\paragraph{Video restoration.}
Here we discuss some work from video restoration as it is related to our idea of restoring low-quality video while doing stereo video generation.
We inspire by Real-ESRGAN~\cite{wang2021real} which propose to train image super-resolution models via degrading synthetic data.
\citet{liang2022recurrent, liang2024vrt} propose video restoration transformers (e.g., RVRT, VRT) for temporally coherent video super-resolution, deblurring, denoising, etc.
DiffBIR~\cite{lin2024diffbir} and DiffIR2VR-Zero~\cite{yeh2024diffir2vr} start to consider using generative diffusion prior to improve generalization capability across various degradation types and datasets.
Upscale-A-Video~\cite{zhou2024upscale} is a diffusion-based video restoration model but requires significant GPU memory to run.

Flow-guided super-resolution models, such as BasicVSR~\cite{chan2021basicvsr}, BasicVSR++~\cite{chan2022basicvsr++} are another type of models that can achieve video restoration efficiently with spatial-temporal consistency.
FMA-Net~\cite{youk2024fma} further advances the quality of joint video super-resolution and deblurring via carefully designed flow-guided modules.
Incorporating these off-the-shelf models for stereo video generation and restoration requires more storage and resources to run. The additional restoration step, without being integrated into the stereo generation pipeline, can introduce artifacts and inconsistency that are more visible in the stereo setup.

\section{Method}

\subsection{Overview}
Given an input left-view video represented by $\inputvideo^{(l)}$ with \NumFrames=16 frames, our goal is to generate both the left-view video $\hat{\inputvideo}^{(l)}$ and right-view video $\hat{\inputvideo}^{(r)}$.
Different from previous methods~\cite{wang2024stereodiffusion,zhao2024stereocrafter}, we target at stereo video generation and restoration where input video can be low-resolution and enhancement is required for the left-view as well.
Therefore, our pipeline consists of training both left-to-right and left-to-left generation branches, as shown in~\cref{fig:pipeline}.
We fine-tune a single video diffusion model to for inpainting and restoration during left-to-right generation, as well as restoration during left-to-left generation.

Note that in this paper, we fine-tune the model from a pretrained model instead of training the model from scratch. 
But as our model is not limited to fine-tuning, we will use training and fine-tuning interchangeably.


\begin{figure*}

\newcommand\scalevalue{0.93} 

\newcommand{\PlotSingleImageBB}[1]{%
    \begin{tikzpicture}[scale=\scalevalue]
        \begin{scope}
            \clip (0,0) rectangle (5,2.5);
            \path[fill overzoom image=figures/#1] (0,0) rectangle (5,2.5);
        \end{scope}
        \draw (0,0) rectangle (5,2.5);
    \end{tikzpicture}%
}

\newcommand{\PlotCroppedImageEnlarged}[2]{%
  \begin{tikzpicture}[scale=\scalevalue-0.04]
    \begin{scope}
      \clip (0,0) rectangle (1.25,1.25);
      \path[fill overzoom image=figures/#1] (0,0) rectangle (1.25,1.25);
    \end{scope}
    \draw[line width=0.7pt, red] (0,0) rectangle (1.25,1.25);
    
    \begin{scope}[yshift=1.35cm]
      \clip (0,0) rectangle (1.25,1.25);
      \path[fill overzoom image=figures/#2] (0,0) rectangle (1.25,1.25);
    \end{scope}
    \draw[line width=0.7pt, blue] (0,1.35) rectangle (1.25,2.6);
  \end{tikzpicture}%
}

    

\small
\hspace*{-5mm}
\begin{tabular}{c@{\;}c@{\;}c@{\;}c@{\;}c@{\;}c@{}}
Input video from~\citet{pixabay2025} 
&
& Ours right (without augmentation)
&
& Ours right (with augmentation)
&
\\

\begin{tikzpicture}[scale=\scalevalue]
\PlotSingleImageBB{results_ours_crop/pixabay_dice2/pixabay_dice2_input_left_00015_with_boxes.png}
\end{tikzpicture}
&
\begin{tikzpicture}[scale=\scalevalue]
\PlotCroppedImageEnlarged{results_ours_crop/pixabay_dice2/pixabay_dice2_input_left_00015_crop1.png}{results_ours_crop/pixabay_dice2/pixabay_dice2_input_left_00015_crop2.png}
\end{tikzpicture}
&
\begin{tikzpicture}[scale=\scalevalue]
\PlotSingleImageBB{results_ours_noaug/pixabay_dice2/pixabay_dice2_ours_0.03_right_00015_with_boxes.png}
\end{tikzpicture}
&
\begin{tikzpicture}[scale=\scalevalue]
\PlotCroppedImageEnlarged{results_ours_noaug/pixabay_dice2/pixabay_dice2_ours_0.03_right_00015_crop1.png}{results_ours_noaug/pixabay_dice2/pixabay_dice2_ours_0.03_right_00015_crop2.png}
\end{tikzpicture}
&
\begin{tikzpicture}[scale=\scalevalue]
\PlotSingleImageBB{results_ours_crop/pixabay_dice2/pixabay_dice2_ours_0.03_right_hist_00015_with_boxes.png}
\end{tikzpicture}
&
\begin{tikzpicture}[scale=\scalevalue]
\PlotCroppedImageEnlarged{results_ours_crop/pixabay_dice2/pixabay_dice2_ours_0.03_right_hist_00015_crop1.png}{results_ours_crop/pixabay_dice2/pixabay_dice2_ours_0.03_right_hist_00015_crop2.png}
\end{tikzpicture}

\\[-0.4mm]
\end{tabular}
\centering
\caption{
Data augmentation with degradations is the key for restoration given low-resolution input.
Our right-view output with augmentation contains sharper details around the edges than the one without augmentation.
Input is from~\citet{pixabay2025} degraded to $320 \times 160$ following~\cref{eq:updown}.
}
\label{fig:ablation_data_augmentation}
\end{figure*}
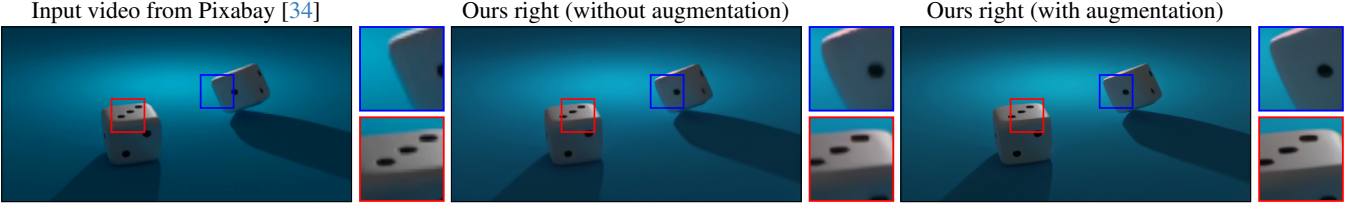

\subsection{Stereo video data generation}
To train such a video diffusion inpaint model, we first need to generate training data.
We use Kubric~\cite{greff2022kubric}, a Blender~\cite{blender2023} based graphics data generation pipeline, to generate synthetic training data in a fully controllable manner.
This allows us to create 3D scenes with customized objects, lighting and camera positions, provided with ground truth data like depth maps.
We rely on the Kubric preprocessed ShapeNet dataset~\cite{chang2015shapenet} and environment map~\cite{hold2019deep} to create realistic scenes.
More specifically, we use a subsets of ShapeNet with 14 classes.
For environment map, we have 458 environment maps preprocessed by Kubric for training.
For each class of object and environment map, we split the dataset to have non-overlapping training and test sets.
For each scene, we generate a left-view video and a right-view video using 2 cameras.
We set the camera baseline to be sampled from a normal distribution with mean 65mm and standard deviation 1mm.
Camera position is randomly sampled on spheres with different radius and the left-camera's look at position is the origin.
The video is generated by moving the objects forward for 21 frames and remove the first 5 frames with remaining $\NumFrames=16$ frames.
We show training examples in~\cref{fig:pipeline} and in supplemental document Fig. 1.


\paragraph{Data augmentation.}
We further augment the dataset by degrading images with a set of blurring, downsampling, adding Gaussian noise and JPEG compression operations, for training the restoration model.
This is inspired by Real-ESRGAN~\cite{wang2021real} to train a generator for image restoration by degrading synthetic images.
To make sure the data degradation is temporally-consistent, we use the implementation of~\citet{wang2021real} while fixing the random seeds during augmentation for all image frames in each video.
We show that data augmentation is crucial for simultaneous generation and restoration in~\cref{fig:ablation_data_augmentation}.
Our generated right-view video with augmentation contains sharper details around the edges than the one without augmentation.


\paragraph{Discussion.}
There exists other large-scale object datasets like Objectverse~\cite{deitke2023objaverse} but they are not preprocessed to be used in Kubric.
Other stereo video datasets like KITTI~\cite{menze2015object} and DrivingStereo~\cite{yang2019drivingstereo} are limited to self-driving scenarios, while Sintel~\cite{butler2012naturalistic} dataset is limited in terms of number of objects and videos.
The StereoCrafter data dataset is not publicly available and requires a lot of resources to capture the data.
Therefore, we believe our dataset will be useful for fine-tuning video diffusion models for stereo video generation in wide range of scenarios.

\subsection{Video warping}
We mainly follow the pipeline of StereoDiffusion~\cite{wang2024stereodiffusion} for disparity-based forward warping.
The forward warping function \forwardwarping performs disparity-based image warping given the left-view video and the depth maps as input.
First, the function computes pixel displacements using the inverse of the depth maps times a disparity scaling \disparityscaling.
Each pixel displaces according to the scaled disparity to generate the forward warped video representing the right-view video.
In some pixel locations of the right-view video, there might be overlapping values and we maintain the pixel value based on the minimum depth utilizing a z-buffer.
For pixels on the right-view that are not assigned any values, we apply bilinear interpolation following StereoCrafter~\cite{zhao2024stereocrafter} to remove those flying pixels.

During training the ground truth depth map is given by the graphics engine, while during inference, we use DepthCrafter~\cite{hu2024depthcrafter} as the video depth estimator. 
According to the depth estimation quality, the warped mask might have unwanted artifacts.
We further postprocess the warping mask using morphological dilation that can eliminate the flying pixels and holes within the occluded regions.
\subsection{Fine-tuning}
\label{sec:training}
Our fine-tuning pipeline consists of both left-to-right and left-to-left generation and restoration.
This is the key difference between previous methods and our proposed method.
Note that in the following, we might use fine-tuning and training interchangeably to present the same meaning.

At each training iteration, we randomly sample a branch as shown in~\cref{fig:pipeline} left-most column.
For the left-to-right branch, input consists of a left-view input video $\inputvideo^{(l)} \in \mathbb{R}^{\NumFrames \times \height \times \width \times 3}$, a warped video $\inputvideo^{(w)} \in \mathbb{R}^{\NumFrames \times \height \times \width \times 3}$ based on disparity-based forward mapping, and a binary warped mask \warpmask representing the occluded region after warping.
$\inputvideo^{(w)}$ is then degraded via our data augmentation to be $\inputvideo'^{(w)}$.
Then $\inputvideo'^{(w)}$ is compressed via a frozen VAE encoder to latents $\latent'^{(w)} \in \mathbb{R}^{\NumFrames \times \height' \times \width' \times 4}$ with a factor of 8 ($\height' = \height / 8, \width' = \width / 8$).
\warpmask is resized to \warpmask' $\in \mathbb{R}^{\NumFrames \times \height' \times \width' \times 1}$, with nearest-neighbor interpolation to ensure that the mask's binary nature is preserved.
As we are training a diffusion model, another input is the target video added with Gaussian noise, represented as $\latent_{t}^{(r)}$, where $t$ is a timestep of a diffusion forward process.
$\latent_{t}^{(r)}$, \warpmask' and $\latent'^{(w)}$ are concatenated along the channel dimension, which serves as the input to the Diffusion U-Net~\cite{blattmann2023stable,ronneberger2015u}.
The output of Diffusion U-Net is $\hat{\latent}^{(r)}$, which is used to compute the mean-square-error (MSE) with $\latent^{(r)}$.
The temporal layers of U-Net is trainable to minimize the MSE loss.

For the left-to-left branch, input becomes the left-view video $\latent^{(l)}$ without warping.
The warped mask becomes a zero mask $\warpmask_0$ indicating that there is no region for inpainting.
After the same data degradation and VAE encoding processes, the input to the U-Net includes a resized warped mask $\warpmask'_0$, a degraded left-view video latent $\latent'^{(l)}$ and a noisy left-view latent $\latent_{t}^{(l)}$.
Similarly, we apply MSE loss between output latent $\hat{\latent}^{(l)}$ and the left-view video latent $\latent^{(l)}$ to optimize the trainable temporal layers in the U-Net.
\subsection{Inference}
\label{sec:inference}
\paragraph{Consistent left and right view generation.}

After training, the model has learnt to maintain or enhance the details in the unmasked regions, as well as to inpaint the mask region, conditioned on the warped mask.
Therefore, we can use $\warpmask$ as an indicator for left-to-right generation and $\warpmask_0$ for left-to-left generation.
During inference, as shown in~\cref{fig:pipeline}, the yellow boxes are no longer used.
We do inference on both left-to-right and left-to-left branches.
After generating the latents $\hat{\latent}^{(l)}$ and $\hat{\latent}^{(r)}$ via iterative diffusion denoising, the VAE decoder is applied (as shown in the green boxes) to reconstruct the videos of both views $\hat{\inputvideo}^{(l)}$ and $\hat{\inputvideo}^{(r)}$.

Note that during inference, we use DepthCrafter~\cite{hu2024depthcrafter} to obtain video depth $D^{(l)}$ and use disparity scaling \disparityscaling as 0.03 for all test videos.
We also show our method is robust to different \disparityscaling values in~\cref{sec:ablation}.

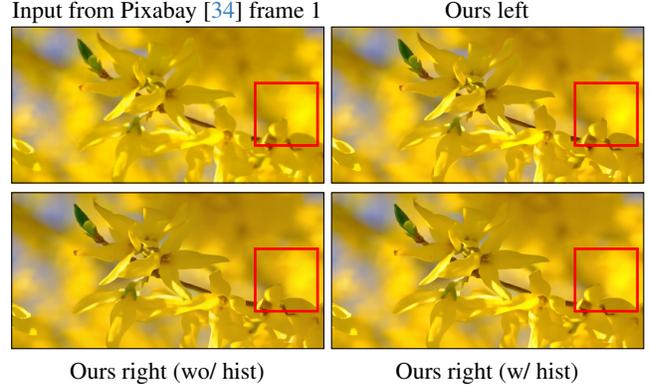
\begin{figure}

        

\newcommand{\PlotSingleImage}[1]{%
    \begin{tikzpicture}[scale=0.83]    
        \begin{scope}
            \clip (0,0) -- (5,0) -- (5,2.5) -- (0,2.5) -- cycle;
            \path[fill overzoom image=figures/#1] (0,0) rectangle (5cm,2.5cm);
        \end{scope}
        \draw (0,0) -- (5,0) -- (5,2.5) -- (0,2.5) -- cycle;
        \draw[line width=1pt, red] (3.9,0.6) rectangle (4.9,1.6);
        
    \end{tikzpicture}%
}

\newcommand\scalevalue{1.2}    
\small
\hspace*{-3.5mm}
\begin{tabular}{c@{\;}c@{}}
Input from Pixabay~\cite{pixabay2025} frame 1
&
Ours left
\\

\begin{tikzpicture}[scale=\scalevalue]
\PlotSingleImage{results_ours/pixabay_flower/pixabay_flower_input_left_00000.png}
\end{tikzpicture}
&
\begin{tikzpicture}[scale=\scalevalue]
\PlotSingleImage{results_ours/pixabay_flower/pixabay_flower_ours_0.03_left_00000.png}
\end{tikzpicture}



\\

\begin{tikzpicture}[scale=\scalevalue]
\PlotSingleImage{results_ours/pixabay_flower/pixabay_flower_ours_0.03_right_00000.png}
\end{tikzpicture}
&
\begin{tikzpicture}[scale=\scalevalue]
\PlotSingleImage{results_ours/pixabay_flower/pixabay_flower_ours_0.03_right_hist_00000.png}
\end{tikzpicture}
\\
Ours right (wo/ hist)
&
Ours right (w/ hist)

\\
[-0.4mm]
\end{tabular}
\centering
\caption{
The color histogram of Ours with histogram matching between left and right are better matched than Ours without histogram matching.
This can be better observed around the red box region where the output gets darker without histogram matching.
Input is from~\citet{pixabay2025} degraded to $320 \times 160$ following~\cref{eq:updown}.
}
\label{fig:histogram_matching}
\end{figure}

\paragraph{Post-processing with histogram matching.}
The output are not guaranteed to have the same brightness, exposure compared to the input.
This is observed in StereoCrafter~\cite{zhao2024stereocrafter}.
To make sure the left and right views are consistent to each other, we post-process the output right-view with input as reference using histogram matching in the scikit-image library~\cite{scikit-image}.
\Cref{fig:histogram_matching} shows that histogram matching can help improving the brightness to better match the input.

\begin{figure*}

\newcommand\scalevalue{0.94} 

\newcommand{\PlotSingleImageBB}[1]{%
    \begin{tikzpicture}[scale=\scalevalue]
        \begin{scope}
            \clip (0,0) rectangle (5,2.5);
            \path[fill overzoom image=figures/#1] (0,0) rectangle (5,2.5);
        \end{scope}
        \draw (0,0) rectangle (5,2.5);
    \end{tikzpicture}%
}

\newcommand{\PlotCroppedImageEnlarged}[2]{%
  \begin{tikzpicture}[scale=\scalevalue-0.04]
    \begin{scope}
      \clip (0,0) rectangle (1.25,1.25);
      \path[fill overzoom image=figures/#1] (0,0) rectangle (1.25,1.25);
    \end{scope}
    \draw[line width=0.7pt, red] (0,0) rectangle (1.25,1.25);
    
    \begin{scope}[yshift=1.35cm]
      \clip (0,0) rectangle (1.25,1.25);
      \path[fill overzoom image=figures/#2] (0,0) rectangle (1.25,1.25);
    \end{scope}
    \draw[line width=0.7pt, blue] (0,1.35) rectangle (1.25,2.6);
  \end{tikzpicture}%
}

    

\small
\hspace*{-8mm}
\begin{tabular}{c@{\;}c@{\;}c@{\;}c@{\;}c@{\;}c@{\;}c@{}}
& Video from~\citet{pixabay2025}
&  
& Video from~\citet{pixabay2025}
& 
& Video from~\citet{pixabay2025}
& 

\\

\rotatebox{90}{\hspace{0.7cm}\small Input}
&
\begin{tikzpicture}[scale=\scalevalue]
\PlotSingleImageBB{results_ours_crop/pixabay_grapes/pixabay_grapes_input_left_00002_with_boxes.png}
\end{tikzpicture}
&
\begin{tikzpicture}[scale=\scalevalue]
\PlotCroppedImageEnlarged{results_ours_crop/pixabay_grapes/pixabay_grapes_input_left_00002_crop1.png}{results_ours_crop/pixabay_grapes/pixabay_grapes_input_left_00002_crop2.png}
\end{tikzpicture}
&
\begin{tikzpicture}[scale=\scalevalue]
\PlotSingleImageBB{results_ours_crop/pixabay_dice1/pixabay_dice1_input_left_00014_with_boxes.png}
\end{tikzpicture}
&
\begin{tikzpicture}[scale=\scalevalue]
\PlotCroppedImageEnlarged{results_ours_crop/pixabay_dice1/pixabay_dice1_input_left_00014_crop1.png}{results_ours_crop/pixabay_dice1/pixabay_dice1_input_left_00014_crop2.png}
\end{tikzpicture}
&
\begin{tikzpicture}[scale=\scalevalue]
\PlotSingleImageBB{results_ours_crop/pixabay_book/pixabay_book_input_left_00005_with_boxes.png}
\end{tikzpicture}
&
\begin{tikzpicture}[scale=\scalevalue]
\PlotCroppedImageEnlarged{results_ours_crop/pixabay_book/pixabay_book_input_left_00005_crop1.png}{results_ours_crop/pixabay_book/pixabay_book_input_left_00005_crop2.png}
\end{tikzpicture}

\\

\rotatebox{90}{\hspace{0.1cm}\small StereoDiffusion}
&
\begin{tikzpicture}[scale=\scalevalue]
\PlotSingleImageBB{results_ours_crop/pixabay_grapes/pixabay_grapes_stereodiffusion_right_00002_with_boxes.png}
\end{tikzpicture}
&
\begin{tikzpicture}[scale=\scalevalue]
\PlotCroppedImageEnlarged{results_ours_crop/pixabay_grapes/pixabay_grapes_stereodiffusion_right_00002_crop1.png}{results_ours_crop/pixabay_grapes/pixabay_grapes_stereodiffusion_right_00002_crop2.png}
\end{tikzpicture}
&
\begin{tikzpicture}[scale=\scalevalue]
\PlotSingleImageBB{results_ours_crop/pixabay_dice1/pixabay_dice1_stereodiffusion_right_00014_with_boxes.png}
\end{tikzpicture}
&
\begin{tikzpicture}[scale=\scalevalue]
\PlotCroppedImageEnlarged{results_ours_crop/pixabay_dice1/pixabay_dice1_stereodiffusion_right_00014_crop1.png}{results_ours_crop/pixabay_dice1/pixabay_dice1_stereodiffusion_right_00014_crop2.png}
\end{tikzpicture}
&
\begin{tikzpicture}[scale=\scalevalue]
\PlotSingleImageBB{results_ours_crop/pixabay_book/pixabay_book_stereodiffusion_right_00005_with_boxes.png}
\end{tikzpicture}
&
\begin{tikzpicture}[scale=\scalevalue]
\PlotCroppedImageEnlarged{results_ours_crop/pixabay_book/pixabay_book_stereodiffusion_right_00005_crop1.png}{results_ours_crop/pixabay_book/pixabay_book_stereodiffusion_right_00005_crop2.png}
\end{tikzpicture}

\\

\rotatebox{90}{\hspace{0.15cm}\small StereoCrafter}
&
\begin{tikzpicture}[scale=\scalevalue]
\PlotSingleImageBB{results_ours_crop/pixabay_grapes/pixabay_grapes_stereocrafter_right_hist_00002_with_boxes.png}
\end{tikzpicture}
&
\begin{tikzpicture}[scale=\scalevalue]
\PlotCroppedImageEnlarged{results_ours_crop/pixabay_grapes/pixabay_grapes_stereocrafter_right_hist_00002_crop1.png}{results_ours_crop/pixabay_grapes/pixabay_grapes_stereocrafter_right_hist_00002_crop2.png}
\end{tikzpicture}
&
\begin{tikzpicture}[scale=\scalevalue]
\PlotSingleImageBB{results_ours_crop/pixabay_dice1/pixabay_dice1_stereocrafter_right_hist_00014_with_boxes.png}
\end{tikzpicture}
&
\begin{tikzpicture}[scale=\scalevalue]
\PlotCroppedImageEnlarged{results_ours_crop/pixabay_dice1/pixabay_dice1_stereocrafter_right_hist_00014_crop1.png}{results_ours_crop/pixabay_dice1/pixabay_dice1_stereocrafter_right_hist_00014_crop2.png}
\end{tikzpicture}
&
\begin{tikzpicture}[scale=\scalevalue]
\PlotSingleImageBB{results_ours_crop/pixabay_book/pixabay_book_stereocrafter_right_hist_00005_with_boxes.png}
\end{tikzpicture}
&
\begin{tikzpicture}[scale=\scalevalue]
\PlotCroppedImageEnlarged{results_ours_crop/pixabay_book/pixabay_book_stereocrafter_right_hist_00005_crop1.png}{results_ours_crop/pixabay_book/pixabay_book_stereocrafter_right_hist_00005_crop2.png}
\end{tikzpicture}

\\

\rotatebox{90}{\hspace{0.5cm}\small Ours left}
&
\begin{tikzpicture}[scale=\scalevalue]
\PlotSingleImageBB{results_ours_crop/pixabay_grapes/pixabay_grapes_ours_0.03_left_00002_with_boxes.png}
\end{tikzpicture}
&
\begin{tikzpicture}[scale=\scalevalue]
\PlotCroppedImageEnlarged{results_ours_crop/pixabay_grapes/pixabay_grapes_ours_0.03_left_00002_crop1.png}{results_ours_crop/pixabay_grapes/pixabay_grapes_ours_0.03_left_00002_crop2.png}
\end{tikzpicture}
&
\begin{tikzpicture}[scale=\scalevalue]
\PlotSingleImageBB{results_ours_crop/pixabay_dice1/pixabay_dice1_ours_0.03_left_00014_with_boxes.png}
\end{tikzpicture}
&
\begin{tikzpicture}[scale=\scalevalue]
\PlotCroppedImageEnlarged{results_ours_crop/pixabay_dice1/pixabay_dice1_ours_0.03_left_00014_crop1.png}{results_ours_crop/pixabay_dice1/pixabay_dice1_ours_0.03_left_00014_crop2.png}
\end{tikzpicture}
&
\begin{tikzpicture}[scale=\scalevalue]
\PlotSingleImageBB{results_ours_crop/pixabay_book/pixabay_book_ours_0.03_left_00005_with_boxes.png}
\end{tikzpicture}
&
\begin{tikzpicture}[scale=\scalevalue]
\PlotCroppedImageEnlarged{results_ours_crop/pixabay_book/pixabay_book_ours_0.03_left_00005_crop1.png}{results_ours_crop/pixabay_book/pixabay_book_ours_0.03_left_00005_crop2.png}
\end{tikzpicture}

\\

\rotatebox{90}{\hspace{0.4cm}\small Ours right}
&
\begin{tikzpicture}[scale=\scalevalue]
\PlotSingleImageBB{results_ours_crop/pixabay_grapes/pixabay_grapes_ours_0.03_right_hist_00002_with_boxes.png}
\end{tikzpicture}
&
\begin{tikzpicture}[scale=\scalevalue]
\PlotCroppedImageEnlarged{results_ours_crop/pixabay_grapes/pixabay_grapes_ours_0.03_right_hist_00002_crop1.png}{results_ours_crop/pixabay_grapes/pixabay_grapes_ours_0.03_right_hist_00002_crop2.png}
\end{tikzpicture}
&
\begin{tikzpicture}[scale=\scalevalue]
\PlotSingleImageBB{results_ours_crop/pixabay_dice1/pixabay_dice1_ours_0.03_right_hist_00014_with_boxes.png}
\end{tikzpicture}
&
\begin{tikzpicture}[scale=\scalevalue]
\PlotCroppedImageEnlarged{results_ours_crop/pixabay_dice1/pixabay_dice1_ours_0.03_right_hist_00014_crop1.png}{results_ours_crop/pixabay_dice1/pixabay_dice1_ours_0.03_right_hist_00014_crop2.png}
\end{tikzpicture}
&
\begin{tikzpicture}[scale=\scalevalue]
\PlotSingleImageBB{results_ours_crop/pixabay_book/pixabay_book_ours_0.03_right_hist_00005_with_boxes.png}
\end{tikzpicture}
&
\begin{tikzpicture}[scale=\scalevalue]
\PlotCroppedImageEnlarged{results_ours_crop/pixabay_book/pixabay_book_ours_0.03_right_hist_00005_crop1.png}{results_ours_crop/pixabay_book/pixabay_book_ours_0.03_right_hist_00005_crop2.png}
\end{tikzpicture}

\\[-0.4mm]
\end{tabular}
\centering
\caption{
Stereo generation comparisons between StereoDiffusion~\cite{wang2024stereodiffusion}, StereoCrafter~\cite{zhao2024stereocrafter} and Ours. 
Our method shows sharper details across different scenes, highlighted in the zoom-in insets.
Inputs are from~\citet{pixabay2025} degraded to $320 \times 160$ following~\cref{eq:updown}.
}
\label{fig:comparison_stereo}
\end{figure*}
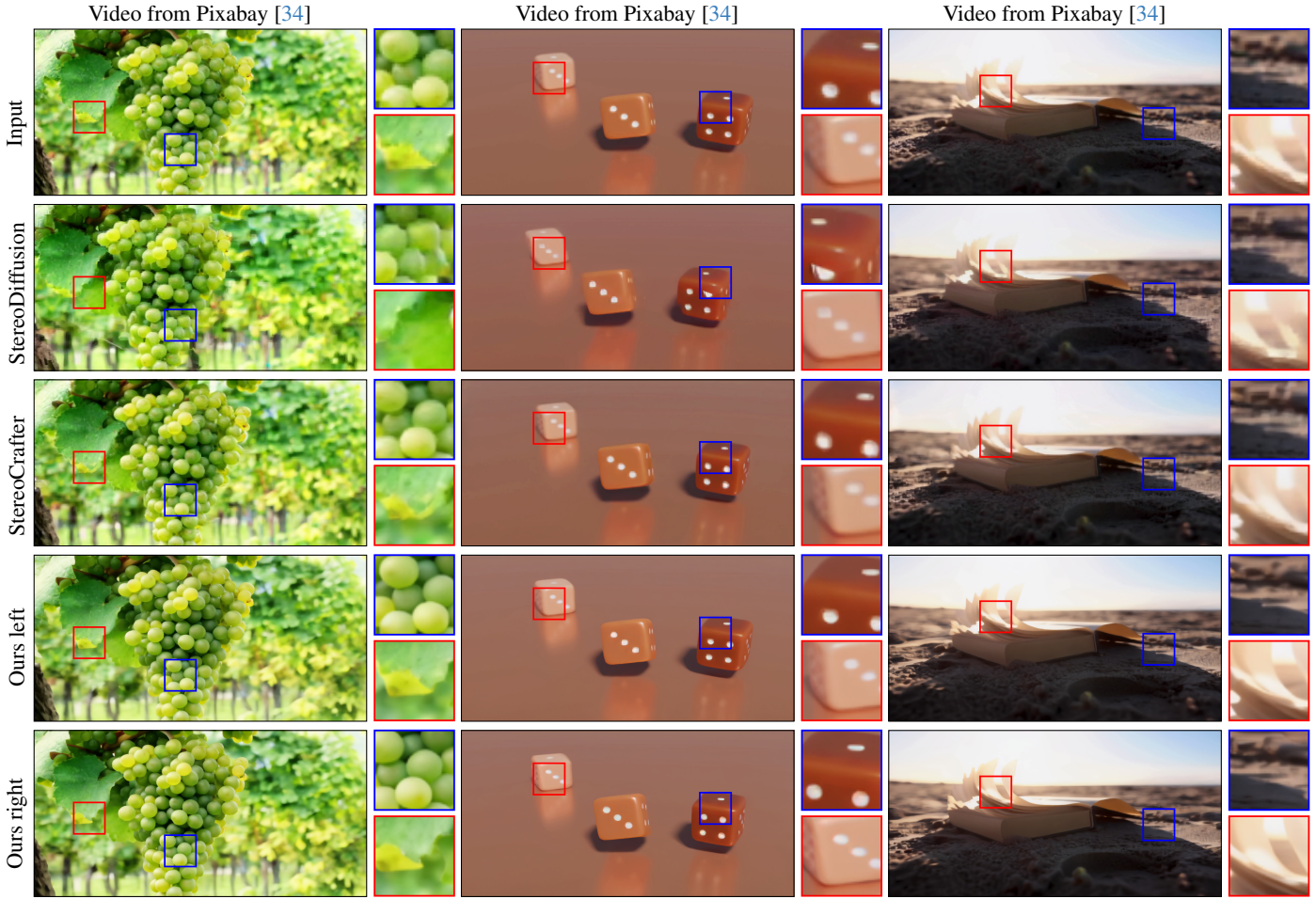

\begin{figure*}
\input{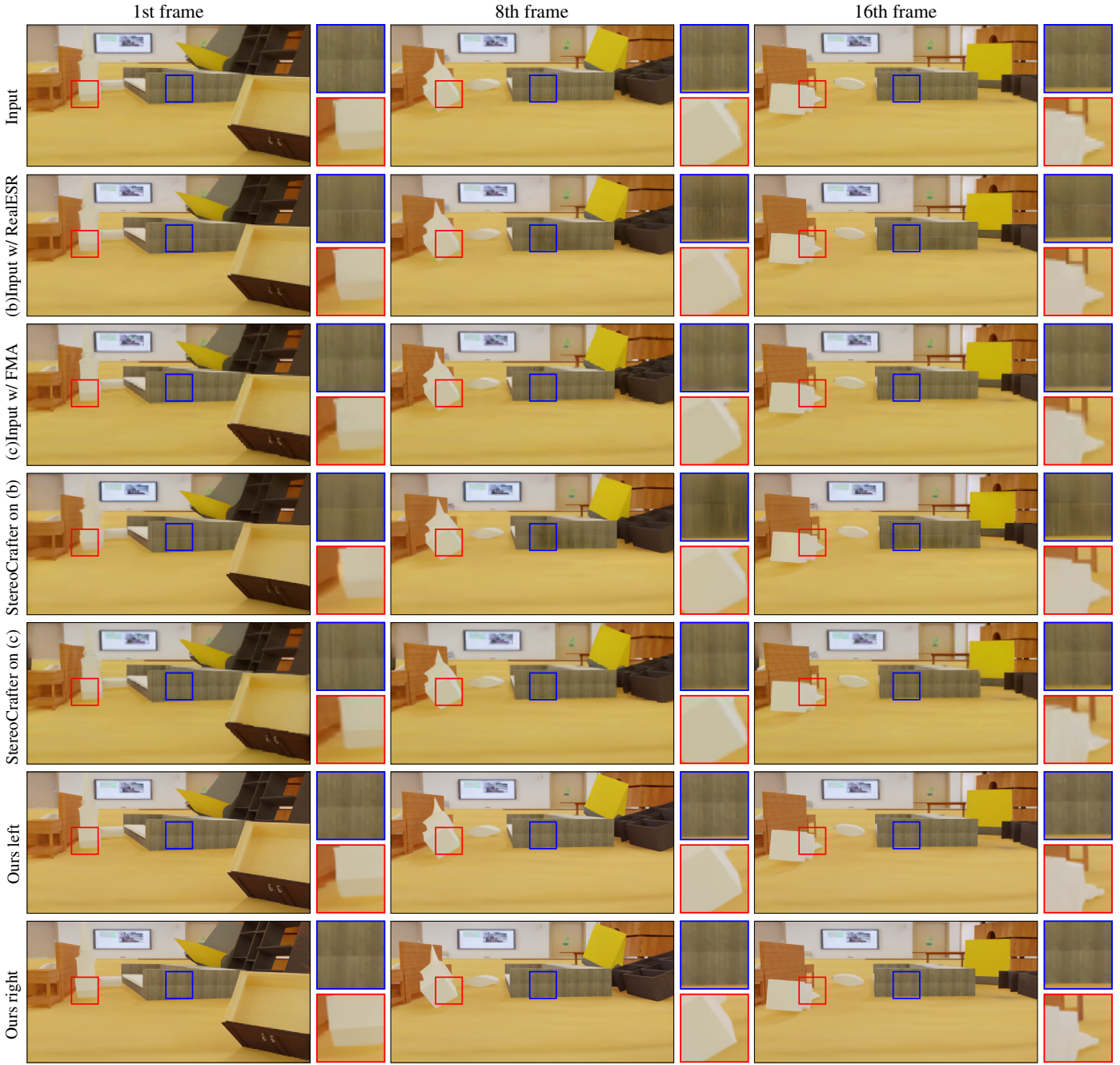}
\centering
\caption{
Stereo generation and restoration comparisons between StereoCrafter~\cite{zhao2024stereocrafter} with FMA-Net~\cite{youk2024fma}, with Real-ESRGAN~\cite{wang2021real} and Ours. 
Our method shows better temporal consistency and image quality than others, highlighted in the zoom-in insets.
Input video is generated in Kubric~\cite{greff2022kubric} degraded to $256 \times 128$ following~\cref{eq:updown}.
}
\label{fig:comparison_stereo_and_restore_kubric_test_02002}
\end{figure*}



\section{Experiments}

\subsection{Implementation details}
\paragraph{Data Generation.}
We preprocess for each input left-view video the warped video and mask via searching a disparity scaling \disparityscaling in [0.02, 0.2], such that the warped video overlaps with the right-view video.
Then we filter out those warped videos that do not overlap with the right-view one.
Finally, we generated 958 videos (each with 16 frames) for training, with a total of 15,328 frames.
For data augmentation, we apply down-sampling, blurring, noising and JPEG compression (following the implementation of~\citet{wang2021real}) with the same random number for each of the video to ensure temporal consistency.
In this way, we double the number of video inputs to 1,916 as training data.

Additionally, we generated 97 test videos using Kubric~\citep{greff2022kubric}.
These test videos are used for evaluating the similarity between generated right-view videos and ground truth right-view videos, using different methods.
For studying user perception, view and temporal consistency, we collect 15 more monocular videos for testing, including  synthetic videos from Kubric~\cite{greff2022kubric}, SVD~\cite{blattmann2023stable}, and 
real-world videos from 
CLEVR~\cite{johnson2017clevr},
Pixabay~\cite{pixabay2025} and some self-captured videos.
All test videos are not unseen during training.

\paragraph{Resolution.}
Training and testing videos are all with spatial resolution $1024 \times 512$.
More specifically, videos are proprocessed with the following equation:
\begin{align}
\label{eq:updown}
    {\inputvideo'} = \texttt{Up}(\texttt{Down}(\inputvideo))
\end{align}
Here we omit the superscript of $\inputvideo$, but $\inputvideo$ and $\inputvideo'$ can represent both left ($\inputvideo^{(l)}$, $\inputvideo'^{(l)}$) and right ($\inputvideo^{(r)}$, $\inputvideo'^{(r)}$) view videos consistent to the notations in~\cref{fig:pipeline}.
Each frame of the video $\inputvideo$ is downsampled to $W' \times H'$ implemented in PyTorch with:
\begin{equation}
\footnotesize
\texttt{Interp} = \texttt{F.interpolate}(x,\; \text{size}=(W',H'),\; \text{mode}=\text{'area'})
\label{eq:interp}
\end{equation}
Note that $\texttt{Down}$ also includes other degraded operations unrelated to resolution like blurring, adding noise and JPEG compression. 

\begin{itemize}
    \item During training/fine-tuning, $W' \times H'=256 \times 128$ or $512 \times 256$ or $1024 \times 512$.
    $\texttt{Up}$ is also implemented as $\texttt{Interp}$ to turn the degraded video back to $1024 \times 512$.
    \item During testing, $\inputvideo_i$ is downsampled ($\texttt{Down}$) to $ 256 \times 128 $, $ 320 \times 160 $ or $ 360 \times 180 $ and then restored back to $1024 \times 512$ via $\texttt{Up} \in $ \{$\texttt{Interp}$, Real-ESRGAN~\citep{wang2021real}, FMA-Net~\citep{youk2024fma}\} to get $\inputvideo'$. 
    $\inputvideo'$ is obtained after restoration and the input to StereoCrafter~\citep{zhao2024stereocrafter}.
\end{itemize}


\paragraph{Fine-tuning.}
We use the Stable Video Diffusion (SVD) architecture
initialized with the trained weights of StereoCrafter.
Following the Pytorch~\cite{paszke2019pytorch} implementation of~\citet{li2023trackdiffusion},
we fine-tune the temporal transformer blocks and use the sampler from~\citet{karras2022elucidating} for training and sampling.
Optimization is performed using AdamW optimizer~\citep{loshchilov2017decoupled} with a learning rate 2e-5 and a batch size of 1.
The resolutions of both input and output videos are $ 1024 \times 512 $, as higher-resolution videos will cause out-of-memory issues.
The degraded videos during training are upsampled back to $ 1024 \times 512 $.
The training takes 10,000 iterations in around 6 hours on an NVIDIA Tesla H100 GPU.

\paragraph{Inference time.}
We test the inference pipeline on an NVIDIA GeForce RTX 4090.
Both our method (Ours) and StereoCrafter~\cite{zhao2024stereocrafter} need to run DepthCrafter~\cite{hu2024depthcrafter} with 20 diffusion sampling steps, which takes around 13 seconds per video (16 frames).
For each video output with 16 frames, the inference of StereoCrafter takes around 19 seconds with 20 diffusion sampling steps.
As we use the same architecture as StereoCrafter and need to run two inferences for generating consistent left-view and right-view videos, our inference time is around 38 seconds.
In total, we take 51 seconds per video while StereoCrafter takes 32 seconds. 
While we can also run StereoCrafter to generate both left-view and right-view videos conditioned on the occlusion mask, we observe that this does not improve the view consistency of StereoCrafter as it is not trained for that.
Histogram matching can be performed in around 5 seconds per output right-view video using the input as reference.



\begin{table*}[ht]
\centering
\small
\caption{Quantitative evaluation on view and frame consistency between 
Ours,
StereoCrafter~\citep{zhao2024stereocrafter} and StereoDiffusion~\citep{wang2024stereodiffusion} from restored input using $\texttt{Interp}$ (\cref{eq:interp}), Real-ESRGAN~\citep{wang2021real}, with FMA-Net~\citep{youk2024fma}.
We show consistent improvements over view and temporal consistency, which is also consistent with the user study in~\Cref{tab:user_study}.
}
\label{tab:quantitative_comparison}
\begin{tabular}{@{}lcccc@{}}
\toprule
Method & LPIPS$_\texttt{view}$($\downarrow$) & LPIPS$_\texttt{temporal}$($\downarrow$) & CLIP$_\texttt{view}$($\uparrow$) & CLIP$_\texttt{temporal}$($\uparrow$) \\ \midrule
StereoCrafter ($\texttt{Up}$=$\texttt{Interp}$) & 0.3052 & 0.1300 & 0.9174 & 0.9854 \\
StereoDiffusion ($\texttt{Up}$=$\texttt{Interp}$) & 0.3419 & 0.1981 & 0.9214 & 0.9736 \\
Stereocrafter ($\texttt{Up}$=Real-ESRGAN) & 0.2551 & 0.1266 & 0.9285 & 0.9852  \\ 
Stereocrafter ($\texttt{Up}$=FMA-Net) & 0.2806 & 0.1327 & 0.9209 & 0.9843  \\ 
Ours & \textbf{0.2393} & \textbf{0.1228} & \textbf{0.9820} & \textbf{0.9874} \\
\bottomrule
\end{tabular}
\end{table*}

\begin{table*}[h]
\centering
\small
\caption{
User study scores on 3D Stereo Effect, Temporal Consistency across frames and Image Quality per frame between 
Ours and StereoCrafter~\cite{zhao2024stereocrafter} with restored input from the $\texttt{Interp}$ and FMA-Net~\cite{youk2024fma} methods.
We show both the average score of all users and scenes, as well as the standard deviation in bracket.
We show consistent improvements across different metrics, which is also consistent to the quantitative metrics in~\Cref{tab:quantitative_comparison}.
}
\label{tab:user_study}
\begin{tabular}{@{}lccc@{}}
\toprule
& StereoCrafter ($\texttt{Up}$=$\texttt{Interp}$) & StereoCrafter ($\texttt{Up}$=FMA-Net) & Ours \\ \midrule
3D Stereo Effect ($\uparrow$) & 3.81 (0.11) & 3.68 (0.16) & \textbf{4.07} (0.10) \\
Temporal Consistency (across frames) ($\uparrow$) & 3.96 (0.05) & 3.81 (0.23) & \textbf{4.28} (0.12) \\
Image Quality (per frame) ($\uparrow$) & 3.17 (0.22) & 3.31 (0.19) & \textbf{4.48} (0.07) \\
\bottomrule
\end{tabular}
\end{table*}

\subsection{Qualitative Comparisons}
We compare our method with two recent diffusion-based methods: StereoDiffusion~\cite{wang2024stereodiffusion}, StereoCrafter~\cite{zhao2024stereocrafter} and its variants.
StereoCrafter is trained to convert any monocular video to a 3D stereoscopic video.
StereoDiffusion is a training-free method to convert an image to a stereo image pair using latent diffusion models~\cite{rombach2022high}, which is not inherently designed for video.
Official code for both methods are available for inference.

As shown in~\cref{fig:comparison_stereo}, both StereoCrafter and StereoDiffusion generate only the right-view video, which preserve most of the details from the input video while doing inpainting on the occluded region.
As the input video is low-resolution, the generated right-view video is also blurry.
This is highlighted in the grapes scene (1st column), the numbers on the dice (2nd column), and the edges of the book pages (3rd column).
Our method improves the image quality for both left and right views.

StereoCrafter can be augmented by existing off-the-shelf video super-resolution method, such as FMA-Net~\cite{youk2024fma} and Real-ESRGAN~\cite{wang2021real}, following~\cref{eq:updown}.
This means the (degraded) input video can be preprocessed by FMA-Net, REAL-ESRGAN or~\texttt{Interp} (if not specified), before being fed into StereoCrafter.
Real-ESRGAN is trained for image super-resolution,
it does not perform temporally consistent generation, as shown in the wood texture (2nd row) of~\cref{fig:comparison_stereo_and_restore_kubric_test_02002}.
FMA-Net is designed for video super-resolution, which takes 3 consecutive frames as input to ensure temporally consistency.
However, it can still introduce inconsistency on the edges of the white object (3rd row).
Further, the upsampling quality of FMA-Net is not as sharp as ours, observed in~\cref{fig:comparison_stereo_and_restore_kubric_test_02002} and supplemental document Fig. 4.
This can be because FMA-Net is designed for joint deblurring and super-resolution on dynamic outdoor scenes, where the architectural design and training objective are different from ours method applied for stereo videos.
Further, we train a single model for both stereo generation and restoration, while StereoCrafter (w/ FMA-Net) requires two independent models, more storage and resources compared to ours.
Our method shows better visual quality in terms of sharpness.
As we take all 16 frames as input, we also work well on temporal consistency.
Lastly, we consider consistent training and inference for both left- and right-view, our generated videos is visually consistent across the two views as shown in~\Cref{fig:comparison_stereo,fig:comparison_stereo_and_restore_kubric_test_02002}.

We also include additional results in supplemental document Figures 3, 4, 5 to show our method improved over others across different scenes, different downsampling scales, and scenes with moving cameras.
More video comparison results can be found in supplemental videos.
Our method can fail to reproduce the details of specular highlight with highly reflected material as shown in supplemental document Fig. 6, as our data for fine-tuning do not contain objects with complex material appearance.

\subsection{Quantitative Comparisons}

\paragraph{Generation quality.}
We use the 97 synthetic test videos rendered with Kubric~\citep{greff2022kubric} to evaluate the right-view generation quality of our method.
We use~\cref{eq:updown} by downsampling left-view input videos $\inputvideo_i$ to $256 \times 128$ and upsampling back to $1024 \times 512$ to obtain $\inputvideo'$.
LPIPS measures perceptual similarity between generated right-view videos and ground truth right-view videos. 
\Cref{tab:lpips_comparison_wrt_gt} shows better performance using our method compared to StereoCrafter~\citep{zhao2024stereocrafter} with different restoration methods.

\begin{table}[H]
\small
\centering
\caption{Comparisons between our method and  StereoCrafter~\citep{zhao2024stereocrafter} with different restoration methods on the quality of right-view video generation.
}
\begin{tabular}{lcc}
\toprule
Method & LPIPS ($\downarrow$) \\
\midrule
StereoCrafter ($\texttt{Up}$=$\texttt{Interp}$) & 0.4271 \\
StereoCrafter ($\texttt{Up}$=FMA-Net~\citep{youk2024fma}) & 0.4232 \\
Ours & \textbf{0.3422} \\
\bottomrule
\end{tabular}
\label{tab:lpips_comparison_wrt_gt}
\end{table}

\paragraph{User study.}
To evaluate human perception, we compare StereoCrafter (\texttt{Up}=\texttt{Interp}), StereoCrafter (\texttt{Up}=FMA-Net) and Ours.
The generated stereo videos of the 3 methods are shown to users using a VR headset (Meta Quest 3).
There are in total 15 videos generated by each of the method, which can be found in the supplemental videos.

We have in total 15 computer science graduate students participating the user study with 3 females and 12 males.
Each user, with normal or corrected vision, was asked to score 5 of the videos generated by all methods based on visual inspection. 
The criterion includes 3D Stereo Effect, Temporal Consistency (across frames) and Image Quality (per frame).
Each user scores from 1 (worst) to 5 (best) to evaluate the quality for each criterion and each method.

The results demonstrate that we achieve clearly the best score on image quality per frame. 
Our results on 3D stereo effect and temporal consistency across frames are also better.

\paragraph{View and temporal consistency.}
We also evaluate the view and temporal consistency using both LPIPS~\cite{zhang2018unreasonable} and CLIP~\cite{radford2021learning} scores, where LPIPS is used to measure the perceptual difference and CLIP is used for semantic similarity.
Following~\citet{dai2024svg}, we use the pretrained CLIP model~\cite{radford2021learning} to extract features for both left and right views of a generated stereo video, and then calculate the feature-wise cosine similarity~\cite{taited2023CLIPScore} to obtain the CLIP view consistency score.
While for temporal consistency score, we compute the feature-wise cosine similarity between the previous frame and current frame.

It is shown in~\Cref{tab:quantitative_comparison} that our method outperforms other methods in terms of view consistency using LPIPS and CLIP.
This can benefit from our design of training and inference with both views in a consistent manner.
While we do not have specific design for temporal consistency, our method still gets slightly better scores.
Given that our evaluation is performed on a relatively small though diverse test set, we do not compute FID~\cite{heusel2017gans} or FVD~\cite{unterthiner2019fvd}. Both are sensitive to the number of videos and requires a large number of reference and generated videos to evaluate.


\subsection{Ablation Study}
\label{sec:ablation}


\paragraph{Data augmentation.}
\Cref{fig:ablation_data_augmentation} demonstrates that our idea of data augmentation with degradations during training is the key to simultaneous stereo video generation and restoration.
As shown in the second column, without data augmentation, the model fails to enhance the quality of the input video.
\Cref{tab:lpips_train_data_size} also shows worse quantitative results by training without data augmentation compared to Ours (training with data degradation).

\paragraph{The effect of training data size.}
Our model performance scales with the increase of data  in terms of both quantity and diversity. 
\Cref{tab:lpips_train_data_size} shows fine-tuning the same model on a smaller-scale dataset of 10 and 100 stereo videos (compared to the original 958), with same epochs but reduced object, motion, scene diversity, leading to lower generation quality.
This indicates the possibility of performance gain over more diverse test videos by fine-tuning with more diverse 3D objects, materials in the dataset.

\begin{table}[H]
\footnotesize
\centering
\caption{Comparisons of training with different datasets on the quality of right-view video generation.
}
\begin{tabular}{lcc}
\toprule
Method & LPIPS ($\downarrow$) \\
\midrule
Ours (fine-tuned wo/ augmentation)                         & 0.3664 \\
Ours (fine-tuned w/ 10 stereo videos) & 0.3878 \\
Ours (fine-tuned w/ 100 stereo videos) & 0.3446 \\
Ours & \textbf{0.3422} \\
\bottomrule
\end{tabular}
\label{tab:lpips_train_data_size}
\end{table}

\paragraph{Performance on high-resolution test videos without degradation.}
\Cref{tab:quantitative_comparison_fullres} shows experiments on the 15 test videos, but with high-resolution ($1024 \times 512$) videos without degradation in~\cref{eq:updown}.
This further emphasizes our fine-tuning with data augmentation works for input videos with varying levels of degradations.

\begin{table}[ht]
\centering
\caption{Quantitative view and frame consistency between StereoCrafter and Ours on high-resolution test videos without degradation.
}
\label{tab:quantitative_comparison_fullres}
\resizebox{\columnwidth}{!}{
\begin{tabular}{@{}lcccc@{}}
\toprule
Method & LPIPS$_{\texttt{view}}$($\downarrow$) & LPIPS$_{\texttt{temporal}}$($\downarrow$) & CLIP$_{\texttt{view}}$($\uparrow$) & CLIP$_{\texttt{temporal}}$($\uparrow$) \\ \midrule
StereoCrafter & 0.2752 & 0.1283 & 0.9317 & 0.9860 \\
Ours & \textbf{0.2695} & \textbf{0.1236} & \textbf{0.9682} & 
\textbf{0.9874} \\
\bottomrule
\end{tabular}
}
\end{table}


\paragraph{Robustness to different disparity scaling \disparityscaling.}
Supplemental document Fig. 2 shows that our generation is robust to different disparity scaling factor during inference, with similar quality output.

\section{Conclusion}
We have presented a novel method that can perform simultaneous stereo video generation and restoration with a single model.
Our method leverages data augmentation with image degradations on synthetic data, as well as a consistent training and inference pipeline to achieve this.
Our designed pipeline enables better results across a diverse set of scenes with low-resolution inputs, compared to existing methods that do not consider both generation and restoration.




{
    \small
    \bibliographystyle{ieeenat_fullname}
    \bibliography{main}

\begin{thebibliography}{56}
\providecommand{\natexlab}[1]{#1}
\providecommand{\url}[1]{\texttt{#1}}
\expandafter\ifx\csname urlstyle\endcsname\relax
  \providecommand{\doi}[1]{doi: #1}\else
  \providecommand{\doi}{doi: \begingroup \urlstyle{rm}\Url}\fi

\bibitem[imm()]{immersityai}
Immersity ai: The ai platform converting images and videos into 3d.
\newblock \url{https://www.immersity.ai/}.

\bibitem[owl()]{owl3d}
Owl3d: Ai-powered 2d to 3d conversion software.
\newblock \url{https://www.owl3d.com/}.

\bibitem[Blattmann et~al.(2023)Blattmann, Dockhorn, Kulal, Mendelevitch, Kilian, Lorenz, Levi, English, Voleti, Letts, et~al.]{blattmann2023stable}
Andreas Blattmann, Tim Dockhorn, Sumith Kulal, Daniel Mendelevitch, Maciej Kilian, Dominik Lorenz, Yam Levi, Zion English, Vikram Voleti, Adam Letts, et~al.
\newblock Stable video diffusion: Scaling latent video diffusion models to large datasets.
\newblock \emph{arXiv preprint arXiv:2311.15127}, 2023.

\bibitem[{Blender Foundation}(2023)]{blender2023}
{Blender Foundation}.
\newblock Blender.
\newblock \url{https://www.blender.org}, 2023.
\newblock Version 3.4.

\bibitem[Butler et~al.(2012)Butler, Wulff, Stanley, and Black]{butler2012naturalistic}
Daniel~J Butler, Jonas Wulff, Garrett~B Stanley, and Michael~J Black.
\newblock A naturalistic open source movie for optical flow evaluation.
\newblock In \emph{Computer Vision--ECCV 2012: 12th European Conference on Computer Vision, Florence, Italy, October 7-13, 2012, Proceedings, Part VI 12}, pages 611--625. Springer, 2012.

\bibitem[Chan et~al.(2021)Chan, Wang, Yu, Dong, and Loy]{chan2021basicvsr}
Kelvin~CK Chan, Xintao Wang, Ke Yu, Chao Dong, and Chen~Change Loy.
\newblock Basicvsr: The search for essential components in video super-resolution and beyond.
\newblock In \emph{Proceedings of the IEEE/CVF conference on computer vision and pattern recognition}, pages 4947--4956, 2021.

\bibitem[Chan et~al.(2022)Chan, Zhou, Xu, and Loy]{chan2022basicvsr++}
Kelvin~CK Chan, Shangchen Zhou, Xiangyu Xu, and Chen~Change Loy.
\newblock Basicvsr++: Improving video super-resolution with enhanced propagation and alignment.
\newblock In \emph{Proceedings of the IEEE/CVF conference on computer vision and pattern recognition}, pages 5972--5981, 2022.

\bibitem[Chang et~al.(2015)Chang, Funkhouser, Guibas, Hanrahan, Huang, Li, Savarese, Savva, Song, Su, et~al.]{chang2015shapenet}
Angel~X Chang, Thomas Funkhouser, Leonidas Guibas, Pat Hanrahan, Qixing Huang, Zimo Li, Silvio Savarese, Manolis Savva, Shuran Song, Hao Su, et~al.
\newblock Shapenet: An information-rich 3d model repository.
\newblock \emph{arXiv preprint arXiv:1512.03012}, 2015.

\bibitem[Dai et~al.(2024)Dai, Tan, Xu, Futschik, Du, Fanello, Qi, and Zhang]{dai2024svg}
Peng Dai, Feitong Tan, Qiangeng Xu, David Futschik, Ruofei Du, Sean Fanello, Xiaojuan Qi, and Yinda Zhang.
\newblock Svg: 3d stereoscopic video generation via denoising frame matrix.
\newblock \emph{arXiv preprint arXiv:2407.00367}, 2024.

\bibitem[Deitke et~al.(2023)Deitke, Schwenk, Salvador, Weihs, Michel, VanderBilt, Schmidt, Ehsani, Kembhavi, and Farhadi]{deitke2023objaverse}
Matt Deitke, Dustin Schwenk, Jordi Salvador, Luca Weihs, Oscar Michel, Eli VanderBilt, Ludwig Schmidt, Kiana Ehsani, Aniruddha Kembhavi, and Ali Farhadi.
\newblock Objaverse: A universe of annotated 3d objects.
\newblock In \emph{Proceedings of the IEEE/CVF conference on computer vision and pattern recognition}, pages 13142--13153, 2023.

\bibitem[Gao et~al.(2024)Gao, Holynski, Henzler, Brussee, Martin-Brualla, Srinivasan, Barron, and Poole]{gao2024cat3d}
Ruiqi Gao, Aleksander Holynski, Philipp Henzler, Arthur Brussee, Ricardo Martin-Brualla, Pratul Srinivasan, Jonathan~T Barron, and Ben Poole.
\newblock Cat3d: Create anything in 3d with multi-view diffusion models.
\newblock \emph{arXiv preprint arXiv:2405.10314}, 2024.

\bibitem[Greff et~al.(2022)Greff, Belletti, Beyer, Doersch, Du, Duckworth, Fleet, Gnanapragasam, Golemo, Herrmann, et~al.]{greff2022kubric}
Klaus Greff, Francois Belletti, Lucas Beyer, Carl Doersch, Yilun Du, Daniel Duckworth, David~J Fleet, Dan Gnanapragasam, Florian Golemo, Charles Herrmann, et~al.
\newblock Kubric: A scalable dataset generator.
\newblock In \emph{Proceedings of the IEEE/CVF conference on computer vision and pattern recognition}, pages 3749--3761, 2022.

\bibitem[Heusel et~al.(2017)Heusel, Ramsauer, Unterthiner, Nessler, and Hochreiter]{heusel2017gans}
Martin Heusel, Hubert Ramsauer, Thomas Unterthiner, Bernhard Nessler, and Sepp Hochreiter.
\newblock Gans trained by a two time-scale update rule converge to a local nash equilibrium.
\newblock \emph{Advances in neural information processing systems}, 30, 2017.

\bibitem[Hold-Geoffroy et~al.(2019)Hold-Geoffroy, Athawale, and Lalonde]{hold2019deep}
Yannick Hold-Geoffroy, Akshaya Athawale, and Jean-Fran{\c{c}}ois Lalonde.
\newblock Deep sky modeling for single image outdoor lighting estimation.
\newblock In \emph{Proceedings of the IEEE/CVF conference on computer vision and pattern recognition}, pages 6927--6935, 2019.

\bibitem[Hou et~al.(2024)Hou, Wei, Zeng, and Chen]{hou2024training}
Chen Hou, Guoqiang Wei, Yan Zeng, and Zhibo Chen.
\newblock Training-free camera control for video generation.
\newblock \emph{arXiv preprint arXiv:2406.10126}, 2024.

\bibitem[Hu et~al.(2024)Hu, Gao, Li, Zhao, Cun, Zhang, Quan, and Shan]{hu2024depthcrafter}
Wenbo Hu, Xiangjun Gao, Xiaoyu Li, Sijie Zhao, Xiaodong Cun, Yong Zhang, Long Quan, and Ying Shan.
\newblock Depthcrafter: Generating consistent long depth sequences for open-world videos.
\newblock \emph{arXiv preprint arXiv:2409.02095}, 2024.

\bibitem[Jin et~al.(2024)Jin, Chen, Liu, Mei, and Zhang]{jin2024t}
Qiao Jin, Xiaodong Chen, Wu Liu, Tao Mei, and Yongdong Zhang.
\newblock T-svg: Text-driven stereoscopic video generation.
\newblock \emph{arXiv preprint arXiv:2412.09323}, 2024.

\bibitem[Johnson et~al.(2017)Johnson, Hariharan, Van Der~Maaten, Fei-Fei, Lawrence~Zitnick, and Girshick]{johnson2017clevr}
Justin Johnson, Bharath Hariharan, Laurens Van Der~Maaten, Li Fei-Fei, C Lawrence~Zitnick, and Ross Girshick.
\newblock Clevr: A diagnostic dataset for compositional language and elementary visual reasoning.
\newblock In \emph{Proceedings of the IEEE conference on computer vision and pattern recognition}, pages 2901--2910, 2017.

\bibitem[Karras et~al.(2022)Karras, Aittala, Aila, and Laine]{karras2022elucidating}
Tero Karras, Miika Aittala, Timo Aila, and Samuli Laine.
\newblock Elucidating the design space of diffusion-based generative models.
\newblock \emph{Advances in neural information processing systems}, 35:\penalty0 26565--26577, 2022.

\bibitem[Kerbl et~al.(2023)Kerbl, Kopanas, Leimk{\"u}hler, and Drettakis]{kerbl20233d}
Bernhard Kerbl, Georgios Kopanas, Thomas Leimk{\"u}hler, and George Drettakis.
\newblock 3d gaussian splatting for real-time radiance field rendering.
\newblock \emph{ACM Trans. Graph.}, 42\penalty0 (4):\penalty0 139--1, 2023.

\bibitem[Li et~al.(2023{\natexlab{a}})Li, Liu, Chen, Hong, Zhuge, Yeung, Lu, and Jia]{li2023trackdiffusion}
Pengxiang Li, Zhili Liu, Kai Chen, Lanqing Hong, Yunzhi Zhuge, Dit-Yan Yeung, Huchuan Lu, and Xu Jia.
\newblock Trackdiffusion: Multi-object tracking data generation via diffusion models.
\newblock \emph{arXiv preprint arXiv:2312.00651}, 2023{\natexlab{a}}.

\bibitem[Li et~al.(2022)Li, Lu, Qin, Guo, and Cheng]{li2022towards}
Zhen Li, Cheng-Ze Lu, Jianhua Qin, Chun-Le Guo, and Ming-Ming Cheng.
\newblock Towards an end-to-end framework for flow-guided video inpainting.
\newblock In \emph{Proceedings of the IEEE/CVF conference on computer vision and pattern recognition}, pages 17562--17571, 2022.

\bibitem[Li et~al.(2023{\natexlab{b}})Li, Wang, Cole, Tucker, and Snavely]{li2023dynibar}
Zhengqi Li, Qianqian Wang, Forrester Cole, Richard Tucker, and Noah Snavely.
\newblock Dynibar: Neural dynamic image-based rendering.
\newblock In \emph{Proceedings of the IEEE/CVF Conference on Computer Vision and Pattern Recognition}, pages 4273--4284, 2023{\natexlab{b}}.

\bibitem[Liang et~al.(2022)Liang, Fan, Xiang, Ranjan, Ilg, Green, Cao, Zhang, Timofte, and Gool]{liang2022recurrent}
Jingyun Liang, Yuchen Fan, Xiaoyu Xiang, Rakesh Ranjan, Eddy Ilg, Simon Green, Jiezhang Cao, Kai Zhang, Radu Timofte, and Luc~V Gool.
\newblock Recurrent video restoration transformer with guided deformable attention.
\newblock \emph{Advances in Neural Information Processing Systems}, 35:\penalty0 378--393, 2022.

\bibitem[Liang et~al.(2024)Liang, Cao, Fan, Zhang, Ranjan, Li, Timofte, and Van~Gool]{liang2024vrt}
Jingyun Liang, Jiezhang Cao, Yuchen Fan, Kai Zhang, Rakesh Ranjan, Yawei Li, Radu Timofte, and Luc Van~Gool.
\newblock Vrt: A video restoration transformer.
\newblock \emph{IEEE Transactions on Image Processing}, 2024.

\bibitem[Lin et~al.(2024)Lin, He, Chen, Lyu, Dai, Yu, Qiao, Ouyang, and Dong]{lin2024diffbir}
Xinqi Lin, Jingwen He, Ziyan Chen, Zhaoyang Lyu, Bo Dai, Fanghua Yu, Yu Qiao, Wanli Ouyang, and Chao Dong.
\newblock Diffbir: Toward blind image restoration with generative diffusion prior.
\newblock In \emph{European Conference on Computer Vision}, pages 430--448. Springer, 2024.

\bibitem[Liu et~al.(2023)Liu, Gao, Meuleman, Tseng, Saraf, Kim, Chuang, Kopf, and Huang]{liu2023robust}
Yu-Lun Liu, Chen Gao, Andreas Meuleman, Hung-Yu Tseng, Ayush Saraf, Changil Kim, Yung-Yu Chuang, Johannes Kopf, and Jia-Bin Huang.
\newblock Robust dynamic radiance fields.
\newblock In \emph{Proceedings of the IEEE/CVF Conference on Computer Vision and Pattern Recognition}, pages 13--23, 2023.

\bibitem[Loshchilov(2017)]{loshchilov2017decoupled}
I Loshchilov.
\newblock Decoupled weight decay regularization.
\newblock \emph{arXiv preprint arXiv:1711.05101}, 2017.

\bibitem[Lugmayr et~al.(2022)Lugmayr, Danelljan, Romero, Yu, Timofte, and Van~Gool]{lugmayr2022repaint}
Andreas Lugmayr, Martin Danelljan, Andres Romero, Fisher Yu, Radu Timofte, and Luc Van~Gool.
\newblock Repaint: Inpainting using denoising diffusion probabilistic models.
\newblock In \emph{Proceedings of the IEEE/CVF conference on computer vision and pattern recognition}, pages 11461--11471, 2022.

\bibitem[Mehl et~al.(2024)Mehl, Bruhn, Gross, and Schroers]{mehl2024stereo}
Lukas Mehl, Andr{\'e}s Bruhn, Markus Gross, and Christopher Schroers.
\newblock Stereo conversion with disparity-aware warping, compositing and inpainting.
\newblock In \emph{Proceedings of the IEEE/CVF Winter Conference on Applications of Computer Vision}, pages 4260--4269, 2024.

\bibitem[Menze and Geiger(2015)]{menze2015object}
Moritz Menze and Andreas Geiger.
\newblock Object scene flow for autonomous vehicles.
\newblock In \emph{Proceedings of the IEEE conference on computer vision and pattern recognition}, pages 3061--3070, 2015.

\bibitem[Mildenhall et~al.(2021)Mildenhall, Srinivasan, Tancik, Barron, Ramamoorthi, and Ng]{mildenhall2021nerf}
Ben Mildenhall, Pratul~P Srinivasan, Matthew Tancik, Jonathan~T Barron, Ravi Ramamoorthi, and Ren Ng.
\newblock Nerf: Representing scenes as neural radiance fields for view synthesis.
\newblock \emph{Communications of the ACM}, 65\penalty0 (1):\penalty0 99--106, 2021.

\bibitem[Paszke et~al.(2019)Paszke, Gross, Massa, Lerer, Bradbury, Chanan, Killeen, Lin, Gimelshein, Antiga, et~al.]{paszke2019pytorch}
Adam Paszke, Sam Gross, Francisco Massa, Adam Lerer, James Bradbury, Gregory Chanan, Trevor Killeen, Zeming Lin, Natalia Gimelshein, Luca Antiga, et~al.
\newblock Pytorch: An imperative style, high-performance deep learning library.
\newblock \emph{Advances in neural information processing systems}, 32, 2019.

\bibitem[{Pixabay}(2025)]{pixabay2025}
{Pixabay}.
\newblock Pixabay license summary.
\newblock \url{https://pixabay.com/service/license-summary/}, 2025.
\newblock Accessed: 2025-03-03.

\bibitem[Radford et~al.(2021)Radford, Kim, Hallacy, Ramesh, Goh, Agarwal, Sastry, Askell, Mishkin, Clark, et~al.]{radford2021learning}
Alec Radford, Jong~Wook Kim, Chris Hallacy, Aditya Ramesh, Gabriel Goh, Sandhini Agarwal, Girish Sastry, Amanda Askell, Pamela Mishkin, Jack Clark, et~al.
\newblock Learning transferable visual models from natural language supervision.
\newblock In \emph{International conference on machine learning}, pages 8748--8763. PmLR, 2021.

\bibitem[Rombach et~al.(2022)Rombach, Blattmann, Lorenz, Esser, and Ommer]{rombach2022high}
Robin Rombach, Andreas Blattmann, Dominik Lorenz, Patrick Esser, and Bj{\"o}rn Ommer.
\newblock High-resolution image synthesis with latent diffusion models.
\newblock In \emph{Proceedings of the IEEE/CVF conference on computer vision and pattern recognition}, pages 10684--10695, 2022.

\bibitem[Ronneberger et~al.(2015)Ronneberger, Fischer, and Brox]{ronneberger2015u}
Olaf Ronneberger, Philipp Fischer, and Thomas Brox.
\newblock U-net: Convolutional networks for biomedical image segmentation.
\newblock In \emph{Medical image computing and computer-assisted intervention--MICCAI 2015: 18th international conference, Munich, Germany, October 5-9, 2015, proceedings, part III 18}, pages 234--241. Springer, 2015.

\bibitem[Shi et~al.(2024{\natexlab{a}})Shi, Li, and Wonka]{shi2024immersepro}
Jian Shi, Zhenyu Li, and Peter Wonka.
\newblock Immersepro: End-to-end stereo video synthesis via implicit disparity learning.
\newblock \emph{arXiv preprint arXiv:2410.00262}, 2024{\natexlab{a}}.

\bibitem[Shi et~al.(2024{\natexlab{b}})Shi, Wang, Li, and Wonka]{shi2024stereocrafterzero}
Jian Shi, Qian Wang, Zhenyu Li, and Peter Wonka.
\newblock Stereocrafter-zero: Zero-shot stereo video generation with noisy restart.
\newblock \emph{arXiv preprint arXiv:2411.14295}, 2024{\natexlab{b}}.

\bibitem[Unterthiner et~al.(2019)Unterthiner, Van~Steenkiste, Kurach, Marinier, Michalski, and Gelly]{unterthiner2019fvd}
Thomas Unterthiner, Sjoerd Van~Steenkiste, Karol Kurach, Rapha{\"e}l Marinier, Marcin Michalski, and Sylvain Gelly.
\newblock Fvd: A new metric for video generation.
\newblock 2019.

\bibitem[van~der Walt et~al.(2014)van~der Walt, {S}ch\"onberger, {Nunez-Iglesias}, {B}oulogne, {W}arner, {Y}ager, {G}ouillart, {Y}u, and the scikit-image contributors]{scikit-image}
{S}t\'efan van~der Walt, {J}ohannes~{L}. {S}ch\"onberger, {J}uan {Nunez-Iglesias}, {F}ran\c{c}ois {B}oulogne, {J}oshua~{D}. {W}arner, {N}eil {Y}ager, {E}mmanuelle {G}ouillart, {T}ony {Y}u, and the scikit-image contributors.
\newblock scikit-image: image processing in {P}ython.
\newblock \emph{PeerJ}, 2:\penalty0 e453, 2014.

\bibitem[Van~Hoorick et~al.(2024)Van~Hoorick, Wu, Ozguroglu, Sargent, Liu, Tokmakov, Dave, Zheng, and Vondrick]{van2024generative}
Basile Van~Hoorick, Rundi Wu, Ege Ozguroglu, Kyle Sargent, Ruoshi Liu, Pavel Tokmakov, Achal Dave, Changxi Zheng, and Carl Vondrick.
\newblock Generative camera dolly: Extreme monocular dynamic novel view synthesis.
\newblock In \emph{European Conference on Computer Vision}, pages 313--331. Springer, 2024.

\bibitem[Wang et~al.(2023)Wang, Yuan, Chen, Zhang, Wang, and Zhang]{wang2023modelscope}
Jiuniu Wang, Hangjie Yuan, Dayou Chen, Yingya Zhang, Xiang Wang, and Shiwei Zhang.
\newblock Modelscope text-to-video technical report.
\newblock \emph{arXiv preprint arXiv:2308.06571}, 2023.

\bibitem[Wang et~al.(2024)Wang, Frisvad, Jensen, and Bigdeli]{wang2024stereodiffusion}
Lezhong Wang, Jeppe~Revall Frisvad, Mark~Bo Jensen, and Siavash~Arjomand Bigdeli.
\newblock Stereodiffusion: Training-free stereo image generation using latent diffusion models.
\newblock In \emph{Proceedings of the IEEE/CVF Conference on Computer Vision and Pattern Recognition}, pages 7416--7425, 2024.

\bibitem[Wang et~al.(2021)Wang, Xie, Dong, and Shan]{wang2021real}
Xintao Wang, Liangbin Xie, Chao Dong, and Ying Shan.
\newblock Real-esrgan: Training real-world blind super-resolution with pure synthetic data.
\newblock In \emph{Proceedings of the IEEE/CVF international conference on computer vision}, pages 1905--1914, 2021.

\bibitem[Wu et~al.(2024)Wu, Gao, Poole, Trevithick, Zheng, Barron, and Holynski]{wu2024cat4d}
Rundi Wu, Ruiqi Gao, Ben Poole, Alex Trevithick, Changxi Zheng, Jonathan~T Barron, and Aleksander Holynski.
\newblock Cat4d: Create anything in 4d with multi-view video diffusion models.
\newblock \emph{arXiv preprint arXiv:2411.18613}, 2024.

\bibitem[Xie et~al.(2024)Xie, Yao, Voleti, Jiang, and Jampani]{xie2024sv4d}
Yiming Xie, Chun-Han Yao, Vikram Voleti, Huaizu Jiang, and Varun Jampani.
\newblock Sv4d: Dynamic 3d content generation with multi-frame and multi-view consistency.
\newblock \emph{arXiv preprint arXiv:2407.17470}, 2024.

\bibitem[Yang et~al.(2019)Yang, Song, Huang, Deng, Shi, and Zhou]{yang2019drivingstereo}
Guorun Yang, Xiao Song, Chaoqin Huang, Zhidong Deng, Jianping Shi, and Bolei Zhou.
\newblock Drivingstereo: A large-scale dataset for stereo matching in autonomous driving scenarios.
\newblock In \emph{Proceedings of the IEEE/CVF conference on computer vision and pattern recognition}, pages 899--908, 2019.

\bibitem[Yeh et~al.(2024)Yeh, Lin, Wang, Hsiao, Chen, Shiu, and Liu]{yeh2024diffir2vr}
Chang-Han Yeh, Chin-Yang Lin, Zhixiang Wang, Chi-Wei Hsiao, Ting-Hsuan Chen, Hau-Shiang Shiu, and Yu-Lun Liu.
\newblock Diffir2vr-zero: Zero-shot video restoration with diffusion-based image restoration models.
\newblock \emph{arXiv preprint arXiv:2407.01519}, 2024.

\bibitem[Youk et~al.(2024)Youk, Oh, and Kim]{youk2024fma}
Geunhyuk Youk, Jihyong Oh, and Munchurl Kim.
\newblock Fma-net: Flow-guided dynamic filtering and iterative feature refinement with multi-attention for joint video super-resolution and deblurring.
\newblock In \emph{Proceedings of the IEEE/CVF Conference on Computer Vision and Pattern Recognition}, pages 44--55, 2024.

\bibitem[Zhang et~al.(2024)Zhang, Jia, Liu, Zhang, Wei, and Tian]{zhang2024spatialme}
Jiale Zhang, Qianxi Jia, Yang Liu, Wei Zhang, Wei Wei, and Xin Tian.
\newblock Spatialme: Stereo video conversion using depth-warping and blend-inpainting.
\newblock \emph{arXiv preprint arXiv:2412.11512}, 2024.

\bibitem[Zhang et~al.(2018)Zhang, Isola, Efros, Shechtman, and Wang]{zhang2018unreasonable}
Richard Zhang, Phillip Isola, Alexei~A Efros, Eli Shechtman, and Oliver Wang.
\newblock The unreasonable effectiveness of deep features as a perceptual metric.
\newblock In \emph{Proceedings of the IEEE conference on computer vision and pattern recognition}, pages 586--595, 2018.

\bibitem[Zhao et~al.(2024)Zhao, Hu, Cun, Zhang, Li, Kong, Gao, Niu, and Shan]{zhao2024stereocrafter}
Sijie Zhao, Wenbo Hu, Xiaodong Cun, Yong Zhang, Xiaoyu Li, Zhe Kong, Xiangjun Gao, Muyao Niu, and Ying Shan.
\newblock Stereocrafter: Diffusion-based generation of long and high-fidelity stereoscopic 3d from monocular videos.
\newblock \emph{arXiv preprint arXiv:2409.07447}, 2024.

\bibitem[Zhengwentai(2023)]{taited2023CLIPScore}
SUN Zhengwentai.
\newblock {clip-score: CLIP Score for PyTorch}.
\newblock \url{https://github.com/taited/clip-score}, 2023.
\newblock Version 0.1.1.

\bibitem[Zhou et~al.(2023)Zhou, Li, Chan, and Loy]{zhou2023propainter}
Shangchen Zhou, Chongyi Li, Kelvin~C.K Chan, and Chen~Change Loy.
\newblock {ProPainter}: Improving propagation and transformer for video inpainting.
\newblock In \emph{Proceedings of IEEE International Conference on Computer Vision (ICCV)}, 2023.

\bibitem[Zhou et~al.(2024)Zhou, Yang, Wang, Luo, and Loy]{zhou2024upscale}
Shangchen Zhou, Peiqing Yang, Jianyi Wang, Yihang Luo, and Chen~Change Loy.
\newblock Upscale-a-video: Temporal-consistent diffusion model for real-world video super-resolution.
\newblock In \emph{Proceedings of the IEEE/CVF Conference on Computer Vision and Pattern Recognition}, pages 2535--2545, 2024.

\end{thebibliography}
}


\end{document}


\maketitle

In the supplemental document, we include additional results.


\section{Additional results}

\paragraph{Data augmentation examples.}
We show in~\cref{fig:training_data} that we apply degradations as data augmentation, including blurring, downsampling, adding Gaussian noise and JPEG compression.
The degradation is applied with the same random seed in a temporally-consistent manner to all the frames of the input left-view video.
\begin{figure*}

\newcommand{\PlotSingleImage}[1]{%
        \begin{scope}
            \clip (0,0) -- (5,0) -- (5,2.5) -- (0,2.5) -- cycle;
            \path[fill overzoom image=figures/#1] (0,0) rectangle (5cm,2.5cm);
        \end{scope}
        \draw (0,0) -- (5,0) -- (5,2.5) -- (0,2.5) -- cycle;
        
}



\newcommand\scalevalue{1.1}    
\small
\begin{tabular}{c@{\;}}
\begin{tabular}{c@{\;}c@{\;}c@{\;}c@{}}
\\[-0.4mm]
~ &
1st frame & 
10th frame &
16th frame
\\
\rotatebox{90}{\hspace{0.3cm} \small Input left-view}
&
\begin{tikzpicture}[scale=\scalevalue]
\PlotSingleImage{training_data/seed00001_elev00017_r00072_azim32_fov97_obj31_base64/left/00005.png}
\end{tikzpicture}
&
\begin{tikzpicture}[scale=\scalevalue]
\PlotSingleImage{training_data/seed00001_elev00017_r00072_azim32_fov97_obj31_base64/left/00014.png}
\end{tikzpicture}
&
\begin{tikzpicture}[scale=\scalevalue]
\PlotSingleImage{training_data/seed00001_elev00017_r00072_azim32_fov97_obj31_base64/left/00020.png}
\end{tikzpicture}
\\
\rotatebox{90}{\hspace{0.1cm} \small Data augmentation}
&
\begin{tikzpicture}[scale=\scalevalue]
\PlotSingleImage{training_data/seed00001_elev00017_r00072_azim32_fov97_obj31_base64/left_aug1/00005.png}
\end{tikzpicture}
&
\begin{tikzpicture}[scale=\scalevalue]
\PlotSingleImage{training_data/seed00001_elev00017_r00072_azim32_fov97_obj31_base64/left_aug1/00014.png}
\end{tikzpicture}
&
\begin{tikzpicture}[scale=\scalevalue]
\PlotSingleImage{training_data/seed00001_elev00017_r00072_azim32_fov97_obj31_base64/left_aug1/00020.png}
\end{tikzpicture}
\\
\rotatebox{90}{\hspace{0.1cm} \small Target right-view}
&
\begin{tikzpicture}[scale=\scalevalue]
\PlotSingleImage{training_data/seed00001_elev00017_r00072_azim32_fov97_obj31_base64/right/00005.png}
\end{tikzpicture}
&
\begin{tikzpicture}[scale=\scalevalue]
\PlotSingleImage{training_data/seed00001_elev00017_r00072_azim32_fov97_obj31_base64/right/00014.png}
\end{tikzpicture}
&
\begin{tikzpicture}[scale=\scalevalue]
\PlotSingleImage{training_data/seed00001_elev00017_r00072_azim32_fov97_obj31_base64/right/00020.png}
\end{tikzpicture}
\\[-0.4mm]
\end{tabular}
\end{tabular}
\centering
\caption{
Training data augmentation with degradation of blurring, downsampling, adding Gaussian noise and JPEG compression in a temporally consistent manner with fixed random seed.
}
\label{fig:training_data}
\end{figure*}

\paragraph{Robustness to different disparity scaling \disparityscaling.}
During inference, our method is robust to different disparity scaling \disparityscaling values.
\Cref{fig:ablation_disparity_supp} shows that our generated left-view and right-view videos have similar quality under different values of \disparityscaling.
\begin{figure*}

        

\newcommand{\PlotSingleImage}[1]{%
    \begin{tikzpicture}[scale=1.1]    
        \begin{scope}
            \clip (0,0) -- (5,0) -- (5,2.5) -- (0,2.5) -- cycle;
            \path[fill overzoom image=figures/#1] (0,0) rectangle (5cm,2.5cm);
        \end{scope}
        \draw (0,0) -- (5,0) -- (5,2.5) -- (0,2.5) -- cycle;
        
    \end{tikzpicture}%
}

\newcommand\scalevalue{1.2}    
\small
\begin{tabular}{c@{\;}c@{\;}c@{\;}c@{}}
& Video from Pixabay~\cite{pixabay2025} (\disparityscaling =  0.03) & \disparityscaling = 0.04 & \disparityscaling = 0.05
\\
\rotatebox{90}{\hspace{0.0cm} \small Input warped videos}
&
\begin{tikzpicture}[scale=\scalevalue]
\PlotSingleImage{results_ours/pixabay_smoke/pixabay_smoke_ours_0.03_warp_00015.png}
\end{tikzpicture}
&
\begin{tikzpicture}[scale=\scalevalue]
\PlotSingleImage{results_ours/pixabay_smoke/pixabay_smoke_ours_0.04_warp_00015.png}
\end{tikzpicture}
&
\begin{tikzpicture}[scale=\scalevalue]
\PlotSingleImage{results_ours/pixabay_smoke/pixabay_smoke_ours_0.05_warp_00015.png}
\end{tikzpicture}
\\
\rotatebox{90}{\hspace{0.5cm} \small Ours left}
&
\begin{tikzpicture}[scale=\scalevalue]
\PlotSingleImage{results_ours/pixabay_smoke/pixabay_smoke_ours_0.03_left_00015.png}
\end{tikzpicture}
&
\begin{tikzpicture}[scale=\scalevalue]
\PlotSingleImage{results_ours/pixabay_smoke/pixabay_smoke_ours_0.04_left_00015.png}
\end{tikzpicture}
&
\begin{tikzpicture}[scale=\scalevalue]
\PlotSingleImage{results_ours/pixabay_smoke/pixabay_smoke_ours_0.05_left_00015.png}
\end{tikzpicture}

\\
\rotatebox{90}{\hspace{0.5cm} \small Ours right}
&
\begin{tikzpicture}[scale=\scalevalue]
\PlotSingleImage{results_ours/pixabay_smoke/pixabay_smoke_ours_0.03_right_hist_00015.png}
\end{tikzpicture}
&
\begin{tikzpicture}[scale=\scalevalue]
\PlotSingleImage{results_ours/pixabay_smoke/pixabay_smoke_ours_0.04_right_hist_00015.png}
\end{tikzpicture}
&
\begin{tikzpicture}[scale=\scalevalue]
\PlotSingleImage{results_ours/pixabay_smoke/pixabay_smoke_ours_0.05_right_hist_00015.png}
\end{tikzpicture}
\\
[-0.4mm]
\end{tabular}
\centering
\caption{
Our method is robust to different disparity scaling values \disparityscaling.
}
\label{fig:ablation_disparity_supp}
\end{figure*}

\paragraph{More comparisons for stereo video generation.}
We show in~\cref{fig:comparison_stereo_supp} more comparisons regarding stereo video generation.
We can observe the same behavior that StereoCrafter~\cite{zhao2024stereocrafter} and StereoDiffusion~\cite{wang2024stereodiffusion} fail to improve the video quality, while our method is designed for both generation and restoration.
Our method shows better visual quality highlighted in the zoom-in insets.

\begin{figure*}

\newcommand\scalevalue{0.94} 

\newcommand{\PlotSingleImageBB}[1]{%
    \begin{tikzpicture}[scale=\scalevalue]
        \begin{scope}
            \clip (0,0) rectangle (5,2.5);
            \path[fill overzoom image=figures/#1] (0,0) rectangle (5,2.5);
        \end{scope}
        \draw (0,0) rectangle (5,2.5);
    \end{tikzpicture}%
}

\newcommand{\PlotCroppedImageEnlarged}[2]{%
  \begin{tikzpicture}[scale=\scalevalue-0.04]
    \begin{scope}
      \clip (0,0) rectangle (1.25,1.25);
      \path[fill overzoom image=figures/#1] (0,0) rectangle (1.25,1.25);
    \end{scope}
    \draw[line width=0.7pt, red] (0,0) rectangle (1.25,1.25);
    
    \begin{scope}[yshift=1.35cm]
      \clip (0,0) rectangle (1.25,1.25);
      \path[fill overzoom image=figures/#2] (0,0) rectangle (1.25,1.25);
    \end{scope}
    \draw[line width=0.7pt, blue] (0,1.35) rectangle (1.25,2.6);
  \end{tikzpicture}%
}

    

\small
\hspace*{-8mm}
\begin{tabular}{c@{\;}c@{\;}c@{\;}c@{\;}c@{\;}c@{\;}c@{}}
& Video from SVD~\cite{rombach2022high} 
&  
& Video from CLEVR~\cite{johnson2017clevr}
& 
& Video from SVD~\cite{rombach2022high}
& 

\\

\rotatebox{90}{\hspace{0.7cm}\small Input}
&
\begin{tikzpicture}[scale=\scalevalue]
\PlotSingleImageBB{results_ours_crop/svd_robot/svd_robot_input_left_00000_with_boxes.png}
\end{tikzpicture}
&
\begin{tikzpicture}[scale=\scalevalue]
\PlotCroppedImageEnlarged{results_ours_crop/svd_robot/svd_robot_input_left_00000_crop1.png}{results_ours_crop/svd_robot/svd_robot_input_left_00000_crop2.png}
\end{tikzpicture}
&
\begin{tikzpicture}[scale=\scalevalue]
\PlotSingleImageBB{results_ours_crop/clever_video_10010/clever_video_10010_input_left_00009_with_boxes.png}
\end{tikzpicture}
&
\begin{tikzpicture}[scale=\scalevalue]
\PlotCroppedImageEnlarged{results_ours_crop/clever_video_10010/clever_video_10010_input_left_00009_crop1.png}{results_ours_crop/clever_video_10010/clever_video_10010_input_left_00009_crop2.png}
\end{tikzpicture}
&
\begin{tikzpicture}[scale=\scalevalue]
\PlotSingleImageBB{results_ours_crop/svd_car/svd_car_input_left_00005_with_boxes.png}
\end{tikzpicture}
&
\begin{tikzpicture}[scale=\scalevalue]
\PlotCroppedImageEnlarged{results_ours_crop/svd_car/svd_car_input_left_00005_crop1.png}{results_ours_crop/svd_car/svd_car_input_left_00005_crop2.png}
\end{tikzpicture}

\\

\rotatebox{90}{\hspace{0.1cm}\small StereoDiffusion}
&
\begin{tikzpicture}[scale=\scalevalue]
\PlotSingleImageBB{results_ours_crop/svd_robot/svd_robot_stereodiffusion_right_00000_with_boxes.png}
\end{tikzpicture}
&
\begin{tikzpicture}[scale=\scalevalue]
\PlotCroppedImageEnlarged{results_ours_crop/svd_robot/svd_robot_stereodiffusion_right_00000_crop1.png}{results_ours_crop/svd_robot/svd_robot_stereodiffusion_right_00000_crop2.png}
\end{tikzpicture}
&
\begin{tikzpicture}[scale=\scalevalue]
\PlotSingleImageBB{results_ours_crop/clever_video_10010/clever_video_10010_stereodiffusion_right_00009_with_boxes.png}
\end{tikzpicture}
&
\begin{tikzpicture}[scale=\scalevalue]
\PlotCroppedImageEnlarged{results_ours_crop/clever_video_10010/clever_video_10010_stereodiffusion_right_00009_crop1.png}{results_ours_crop/clever_video_10010/clever_video_10010_stereodiffusion_right_00009_crop2.png}
\end{tikzpicture}
&
\begin{tikzpicture}[scale=\scalevalue]
\PlotSingleImageBB{results_ours_crop/svd_car/svd_car_stereodiffusion_right_00005_with_boxes.png}
\end{tikzpicture}
&
\begin{tikzpicture}[scale=\scalevalue]
\PlotCroppedImageEnlarged{results_ours_crop/svd_car/svd_car_stereodiffusion_right_00005_crop1.png}{results_ours_crop/svd_car/svd_car_stereodiffusion_right_00005_crop2.png}
\end{tikzpicture}

\\

\rotatebox{90}{\hspace{0.15cm}\small StereoCrafter}
&
\begin{tikzpicture}[scale=\scalevalue]
\PlotSingleImageBB{results_ours_crop/svd_robot/svd_robot_stereocrafter_right_hist_00000_with_boxes.png}
\end{tikzpicture}
&
\begin{tikzpicture}[scale=\scalevalue]
\PlotCroppedImageEnlarged{results_ours_crop/svd_robot/svd_robot_stereocrafter_right_hist_00000_crop1.png}{results_ours_crop/svd_robot/svd_robot_stereocrafter_right_hist_00000_crop2.png}
\end{tikzpicture}
&
\begin{tikzpicture}[scale=\scalevalue]
\PlotSingleImageBB{results_ours_crop/clever_video_10010/clever_video_10010_stereocrafter_right_hist_00009_with_boxes.png}
\end{tikzpicture}
&
\begin{tikzpicture}[scale=\scalevalue]
\PlotCroppedImageEnlarged{results_ours_crop/clever_video_10010/clever_video_10010_stereocrafter_right_hist_00009_crop1.png}{results_ours_crop/clever_video_10010/clever_video_10010_stereocrafter_right_hist_00009_crop2.png}
\end{tikzpicture}
&
\begin{tikzpicture}[scale=\scalevalue]
\PlotSingleImageBB{results_ours_crop/svd_car/svd_car_stereocrafter_right_hist_00005_with_boxes.png}
\end{tikzpicture}
&
\begin{tikzpicture}[scale=\scalevalue]
\PlotCroppedImageEnlarged{results_ours_crop/svd_car/svd_car_stereocrafter_right_hist_00005_crop1.png}{results_ours_crop/svd_car/svd_car_stereocrafter_right_hist_00005_crop2.png}
\end{tikzpicture}

\\

\rotatebox{90}{\hspace{0.5cm}\small Ours left}
&
\begin{tikzpicture}[scale=\scalevalue]
\PlotSingleImageBB{results_ours_crop/svd_robot/svd_robot_ours_0.03_left_00000_with_boxes.png}
\end{tikzpicture}
&
\begin{tikzpicture}[scale=\scalevalue]
\PlotCroppedImageEnlarged{results_ours_crop/svd_robot/svd_robot_ours_0.03_left_00000_crop1.png}{results_ours_crop/svd_robot/svd_robot_ours_0.03_left_00000_crop2.png}
\end{tikzpicture}
&
\begin{tikzpicture}[scale=\scalevalue]
\PlotSingleImageBB{results_ours_crop/clever_video_10010/clever_video_10010_ours_0.03_left_00009_with_boxes.png}
\end{tikzpicture}
&
\begin{tikzpicture}[scale=\scalevalue]
\PlotCroppedImageEnlarged{results_ours_crop/clever_video_10010/clever_video_10010_ours_0.03_left_00009_crop1.png}{results_ours_crop/clever_video_10010/clever_video_10010_ours_0.03_left_00009_crop2.png}
\end{tikzpicture}
&
\begin{tikzpicture}[scale=\scalevalue]
\PlotSingleImageBB{results_ours_crop/svd_car/svd_car_ours_0.03_left_00005_with_boxes.png}
\end{tikzpicture}
&
\begin{tikzpicture}[scale=\scalevalue]
\PlotCroppedImageEnlarged{results_ours_crop/svd_car/svd_car_ours_0.03_left_00005_crop1.png}{results_ours_crop/svd_car/svd_car_ours_0.03_left_00005_crop2.png}
\end{tikzpicture}

\\

\rotatebox{90}{\hspace{0.4cm}\small Ours right}
&
\begin{tikzpicture}[scale=\scalevalue]
\PlotSingleImageBB{results_ours_crop/svd_robot/svd_robot_ours_0.03_right_hist_00000_with_boxes.png}
\end{tikzpicture}
&
\begin{tikzpicture}[scale=\scalevalue]
\PlotCroppedImageEnlarged{results_ours_crop/svd_robot/svd_robot_ours_0.03_right_hist_00000_crop1.png}{results_ours_crop/svd_robot/svd_robot_ours_0.03_right_hist_00000_crop2.png}
\end{tikzpicture}
&
\begin{tikzpicture}[scale=\scalevalue]
\PlotSingleImageBB{results_ours_crop/clever_video_10010/clever_video_10010_ours_0.03_right_hist_00009_with_boxes.png}
\end{tikzpicture}
&
\begin{tikzpicture}[scale=\scalevalue]
\PlotCroppedImageEnlarged{results_ours_crop/clever_video_10010/clever_video_10010_ours_0.03_right_hist_00009_crop1.png}{results_ours_crop/clever_video_10010/clever_video_10010_ours_0.03_right_hist_00009_crop2.png}
\end{tikzpicture}
&
\begin{tikzpicture}[scale=\scalevalue]
\PlotSingleImageBB{results_ours_crop/svd_car/svd_car_ours_0.03_right_hist_00005_with_boxes.png}
\end{tikzpicture}
&
\begin{tikzpicture}[scale=\scalevalue]
\PlotCroppedImageEnlarged{results_ours_crop/svd_car/svd_car_ours_0.03_right_hist_00005_crop1.png}{results_ours_crop/svd_car/svd_car_ours_0.03_right_hist_00005_crop2.png}
\end{tikzpicture}

\\[-0.4mm]
\end{tabular}
\centering
\caption{
Stereo generation comparisons between StereoDiffusion~\cite{wang2024stereodiffusion}, StereoCrafter~\cite{zhao2024stereocrafter} and Ours. 
We show that our method clearly outperforms other methods visually in terms of image quality highlighted in the zoom-in insets.
Input videos are from SVD~\cite{blattmann2023stable} and CLEVR~\cite{johnson2017clevr} downsampled to $256 \times 128$, $360 \times 180$, $360 \times 180$, respectively.
}
\label{fig:comparison_stereo_supp}
\end{figure*}

\paragraph{More comparisons for stereo video generation and restoration.}
\Cref{fig:comparison_stereo_and_restore_pixabay_teapot} shows more comparisons regarding stereo video generation and restoration.
In this scene, both Real-ESRGAN~\cite{wang2021real} and FMA-Net~\cite{youk2024fma} struggle to preserve temporal consistency, e.g., on the textures cropped in the red boxes.
Our method also generates sharper edge details around the cloth than other methods.
\begin{figure*}
\input{figures/comparison_stereo_and_restore_pixabay_teapot}
\centering
\caption{
Stereo generation and restoration comparisons between StereoCrafter with FMA-Net~\cite{youk2024fma}, with Real-ESRGAN~\cite{wang2021real} and Ours. 
We highlight the zoom-in insets that our method generated more temporally-consistent results around the moving textures and edge of the cloth.
The input video is from~\citet{pixabay2025} downsampled to $320 \times 160$.
}
\label{fig:comparison_stereo_and_restore_pixabay_teapot}
\end{figure*}

\paragraph{Moving camera.}
Though our fine-tuning data does not contain scenes with large camera motion, due to the pretrained network, our method can still generate consistent results for scenes with both object and camera motions, as shown in~\cref{fig:comparison_stereo_and_restore_pixabay_turtle1}.
\begin{figure*}

\newcommand\scalevalue{0.94} 

\newcommand{\PlotSingleImageBB}[1]{%
    \begin{tikzpicture}[scale=\scalevalue]
        \begin{scope}
            \clip (0,0) rectangle (5,2.5);
            \path[fill overzoom image=figures/#1] (0,0) rectangle (5,2.5);
        \end{scope}
        \draw (0,0) rectangle (5,2.5);
    \end{tikzpicture}%
}

\newcommand{\PlotCroppedImageEnlarged}[2]{%
  \begin{tikzpicture}[scale=\scalevalue-0.04]
    \begin{scope}
      \clip (0,0) rectangle (1.25,1.25);
      \path[fill overzoom image=figures/#1] (0,0) rectangle (1.25,1.25);
    \end{scope}
    \draw[line width=0.7pt, red] (0,0) rectangle (1.25,1.25);
    
    \begin{scope}[yshift=1.35cm]
      \clip (0,0) rectangle (1.25,1.25);
      \path[fill overzoom image=figures/#2] (0,0) rectangle (1.25,1.25);
    \end{scope}
    \draw[line width=0.7pt, blue] (0,1.35) rectangle (1.25,2.6);
  \end{tikzpicture}%
}

    

\small
\hspace*{-8mm}
\begin{tabular}{c@{\;}c@{\;}c@{\;}c@{\;}c@{\;}c@{\;}c@{}}
& Video from Pixabay~\cite{pixabay2025}: 1st frame 
&  
& 8th frame
& 
& 15th frame
& 

\\

\rotatebox{90}{\hspace{0.7cm}\footnotesize Input}
&
\begin{tikzpicture}[scale=\scalevalue]
\PlotSingleImageBB{results_ours_crop/pixabay_turtle1/pixabay_turtle1_input_left_00000_with_boxes.png}
\end{tikzpicture}
&
\begin{tikzpicture}[scale=\scalevalue]
\PlotCroppedImageEnlarged{results_ours_crop/pixabay_turtle1/pixabay_turtle1_input_left_00000_crop1.png}{results_ours_crop/pixabay_turtle1/pixabay_turtle1_input_left_00000_crop2.png}
\end{tikzpicture}
&
\begin{tikzpicture}[scale=\scalevalue]
\PlotSingleImageBB{results_ours_crop/pixabay_turtle1/pixabay_turtle1_input_left_00007_with_boxes.png}
\end{tikzpicture}
&
\begin{tikzpicture}[scale=\scalevalue]
\PlotCroppedImageEnlarged{results_ours_crop/pixabay_turtle1/pixabay_turtle1_input_left_00007_crop1.png}{results_ours_crop/pixabay_turtle1/pixabay_turtle1_input_left_00007_crop2.png}
\end{tikzpicture}
&
\begin{tikzpicture}[scale=\scalevalue]
\PlotSingleImageBB{results_ours_crop/pixabay_turtle1/pixabay_turtle1_input_left_00014_with_boxes.png}
\end{tikzpicture}
&
\begin{tikzpicture}[scale=\scalevalue]
\PlotCroppedImageEnlarged{results_ours_crop/pixabay_turtle1/pixabay_turtle1_input_left_00014_crop1.png}{results_ours_crop/pixabay_turtle1/pixabay_turtle1_input_left_00014_crop2.png}
\end{tikzpicture}

\\

\rotatebox{90}{\hspace{0.5cm}\footnotesize Ours left}
&
\begin{tikzpicture}[scale=\scalevalue]
\PlotSingleImageBB{results_ours_crop/pixabay_turtle1/pixabay_turtle1_ours_0.03_left_00000_with_boxes.png}
\end{tikzpicture}
&
\begin{tikzpicture}[scale=\scalevalue]
\PlotCroppedImageEnlarged{results_ours_crop/pixabay_turtle1/pixabay_turtle1_ours_0.03_left_00000_crop1.png}{results_ours_crop/pixabay_turtle1/pixabay_turtle1_ours_0.03_left_00000_crop2.png}
\end{tikzpicture}
&
\begin{tikzpicture}[scale=\scalevalue]
\PlotSingleImageBB{results_ours_crop/pixabay_turtle1/pixabay_turtle1_ours_0.03_left_00007_with_boxes.png}
\end{tikzpicture}
&
\begin{tikzpicture}[scale=\scalevalue]
\PlotCroppedImageEnlarged{results_ours_crop/pixabay_turtle1/pixabay_turtle1_ours_0.03_left_00007_crop1.png}{results_ours_crop/pixabay_turtle1/pixabay_turtle1_ours_0.03_left_00007_crop2.png}
\end{tikzpicture}
&
\begin{tikzpicture}[scale=\scalevalue]
\PlotSingleImageBB{results_ours_crop/pixabay_turtle1/pixabay_turtle1_ours_0.03_left_00014_with_boxes.png}
\end{tikzpicture}
&
\begin{tikzpicture}[scale=\scalevalue]
\PlotCroppedImageEnlarged{results_ours_crop/pixabay_turtle1/pixabay_turtle1_ours_0.03_left_00014_crop1.png}{results_ours_crop/pixabay_turtle1/pixabay_turtle1_ours_0.03_left_00014_crop2.png}
\end{tikzpicture}

\\

\rotatebox{90}{\hspace{0.4cm}\footnotesize Ours right}
&
\begin{tikzpicture}[scale=\scalevalue]
\PlotSingleImageBB{results_ours_crop/pixabay_turtle1/pixabay_turtle1_ours_0.03_right_hist_00000_with_boxes.png}
\end{tikzpicture}
&
\begin{tikzpicture}[scale=\scalevalue]
\PlotCroppedImageEnlarged{results_ours_crop/pixabay_turtle1/pixabay_turtle1_ours_0.03_right_hist_00000_crop1.png}{results_ours_crop/pixabay_turtle1/pixabay_turtle1_ours_0.03_right_hist_00000_crop2.png}
\end{tikzpicture}
&
\begin{tikzpicture}[scale=\scalevalue]
\PlotSingleImageBB{results_ours_crop/pixabay_turtle1/pixabay_turtle1_ours_0.03_right_hist_00007_with_boxes.png}
\end{tikzpicture}
&
\begin{tikzpicture}[scale=\scalevalue]
\PlotCroppedImageEnlarged{results_ours_crop/pixabay_turtle1/pixabay_turtle1_ours_0.03_right_hist_00007_crop1.png}{results_ours_crop/pixabay_turtle1/pixabay_turtle1_ours_0.03_right_hist_00007_crop2.png}
\end{tikzpicture}
&
\begin{tikzpicture}[scale=\scalevalue]
\PlotSingleImageBB{results_ours_crop/pixabay_turtle1/pixabay_turtle1_ours_0.03_right_hist_00014_with_boxes.png}
\end{tikzpicture}
&
\begin{tikzpicture}[scale=\scalevalue]
\PlotCroppedImageEnlarged{results_ours_crop/pixabay_turtle1/pixabay_turtle1_ours_0.03_right_hist_00014_crop1.png}{results_ours_crop/pixabay_turtle1/pixabay_turtle1_ours_0.03_right_hist_00014_crop2.png}
\end{tikzpicture}

\\[-0.4mm]
\end{tabular}
\centering
\caption{
This example shows that our method can also generate consistent results for video with both object and camera movements, though each of the training video is generated with a fixed stereo camera position.
The input video is from~\citet{pixabay2025} downsampled to $320 \times 160$.
}
\label{fig:comparison_stereo_and_restore_pixabay_turtle1}
\end{figure*}

\paragraph{Failure case.}
As our dataset only contains simple shapes and materials from the ShapeNet~\cite{chang2015shapenet} dataset.
Our trained model does not generalize well to scenes with highly detailed geometry or appearance, such as the example in~\cref{fig:failure}.
We believe this can be further addressed by considering more complex geometry and material in our dataset.
\begin{figure*}

        

\newcommand{\PlotSingleImage}[1]{%
    \begin{tikzpicture}[scale=1.7]    
        \begin{scope}
            \clip (0,0) -- (5,0) -- (5,2.5) -- (0,2.5) -- cycle;
            \path[fill overzoom image=figures/#1] (0,0) rectangle (5cm,2.5cm);
        \end{scope}
        \draw (0,0) -- (5,0) -- (5,2.5) -- (0,2.5) -- cycle;
        
    \end{tikzpicture}%
}

\newcommand\scalevalue{1.2}    
\small
\begin{tabular}{c@{\;}c@{}}
Input video from CLEVR~\cite{johnson2017clevr}
& 
StereoCrafter~\cite{zhao2024stereocrafter}
\\
\begin{tikzpicture}[scale=\scalevalue]
\PlotSingleImage{results_ours/clever_video_10005/clever_video_10005_input_left_00008.png}
\end{tikzpicture}
& 
\begin{tikzpicture}[scale=\scalevalue]
\PlotSingleImage{results_stereocrafter/clever_video_10005/clever_video_10005_stereocrafter_right_hist_00008.png}
\end{tikzpicture}
\\

\begin{tikzpicture}[scale=\scalevalue]
\PlotSingleImage{results_ours/clever_video_10005/clever_video_10005_ours_0.03_left_00008.png}
\end{tikzpicture}
&
\begin{tikzpicture}[scale=\scalevalue]
\PlotSingleImage{results_ours/clever_video_10005/clever_video_10005_ours_0.03_right_hist_00008.png}
\end{tikzpicture}
\\
Ours left
& 
Ours right
\\
[-0.4mm]
\end{tabular}
\centering
\caption{
Our method can fail to reproduce the details of specular highlight with highly reflected material such as the ball in the scene, as our data for fine-tuning do not contain objects with complex material appearance.
The input video is from~\citet{johnson2017clevr} downsampled to $360 \times 180$.
}
\label{fig:failure}
\end{figure*}

{
    \small
    \bibliographystyle{ieeenat_fullname}
    \bibliography{main}
}
